\documentclass[pdflatex,sn-mathphys-num]{sn-jnl}

\usepackage{graphicx}%
\usepackage{multirow}%
\usepackage{amsmath,amssymb,amsfonts}%
\usepackage{amsthm}%
\usepackage{mathrsfs}%
\usepackage[title]{appendix}%
\usepackage{xcolor}%
\usepackage{textcomp}%
\usepackage{manyfoot}%
\usepackage{booktabs}%
\usepackage{algorithm}%
\usepackage{algorithmicx}%
\usepackage{algpseudocode}%
\usepackage{listings}%
\usepackage{caption}
\usepackage{soul}
\usepackage[T1]{fontenc}
\usepackage{hyperref} 

\theoremstyle{thmstyleone}%
\theoremstyle{thmstyletwo}%

\theoremstyle{thmstylethree}%

\begin{document}

\title[Article Title]{Towards Generalisable Foundation Models for Brain MRI}


\author*[1,2]{\fnm{Moona} \sur{Mazher}}\email{m.mazher@ucl.ac.uk}

\author[2,3]{\fnm{Geoff J. M.} \sur{ Parker}}

\author[1,2]{\fnm{Daniel C.} \sur{Alexander}}

\affil*[1]{\orgdiv{Hawkes Institute, Department of Computer Science}, \orgname{University College London}, \orgaddress{\city{London}, \country{UK}}}

\affil[2]{\orgdiv{Department of Computer Science}, \orgname{University College London}, \orgaddress{\city{London}, \country{UK}}}

\affil[3]{\orgdiv{Department of Medical Physics and Biomedical Engineering}, \orgname{ University College London}, \orgaddress{\city{London}, \country{UK}}}


\abstract{Foundation models in artificial intelligence (AI) are transforming medical imaging by enabling general-purpose feature learning from large-scale, unlabeled datasets. In this work, we introduce BrainFound, a self-supervised foundation model for brain MRI, built by extending DINOv2, a vision transformer originally designed for 2D natural images. BrainFound adapts DINOv2 to model full 3D brain anatomy by incorporating volumetric information from sequential MRI slices, moving beyond conventional single-slice paradigms. It supports both single- and multimodal inputs, enabling a broad range of downstream tasks, including disease detection and image segmentation, while generalising across varied imaging protocols and clinical scenarios. We show that BrainFound consistently outperforms existing self-supervised pretraining strategies and supervised baselines, particularly in label-scarce and multi-contrast settings. By integrating information from diverse 3D MRI modalities (e.g., T1, T2, FLAIR), it enhances diagnostic accuracy and reduces dependency on extensive expert annotations. This flexibility makes BrainFound a scalable and practical solution for 3D neuroimaging pipelines, with significant potential for clinical deployment and research innovation.}




\maketitle

\section{Introduction}\label{sec1}

Radiological imaging is central to modern medicine, offering detailed, non-invasive insights for diagnosis and treatment planning \cite{hussain2022modern, loginoff2023advancements}. With rising imaging use and a shortage of radiologists, the clinical workload has become unsustainable, requiring radiologists to interpret one image every 3-4 seconds to keep up \cite{yoon2024use,kwee2021workload}. This burden leads to delays and diagnostic errors \cite{kasalak2023work,pesapane2024errors}, highlighting the need for automated tools to enhance accuracy and efficiency.

Deep learning has shown promise in radiology, but its success depends on large, labeled datasets, which are expensive and time-consuming to obtain \cite{litjens2017survey,galbusera2024image}. Additionally, supervised models often struggle to generalise across domains and data shifts \cite{yoon2024domain}. Self-supervised learning (SSL) addresses these limitations by learning from unlabeled data, enabling scalable and robust model development \cite{zhang2023dive}.

Foundation Models (FMs), trained using SSL on large-scale data, have achieved strong generalisation and few-shot performance across a diverse range of tasks \cite{schneider2024foundation,brown2020language,hosseinzadeh2023towards}. However, most current models are specifically designed to deal with 2D natural imaging datasets, while medical imaging datasets require 3D models to handle volumetric MRIs \cite{shivdeo2021evaluation}, which are crucial to avoid the loss of spatial context critical for accurate diagnosis.

3D models can capture spatial dependencies across slices, which is essential for tasks such as lesion detection in conditions like multiple sclerosis or neuro-oncology, as well as for assessing brain atrophy in dementia. Working directly with 3D data helps preserve anatomical details. Although the majority of existing machine learning models operate on single-contrast MRI data, our use of multimodal/multi-contrast inputs (e.g., T1, T2, FLAIR) improves performance across a range of tasks, including disease detection and segmentation, by capturing complementary structural information \cite{ferraro2017multimodal,weiner2013alzheimer}.

Self-supervised learning (SSL) has shown promise in brain-related tasks such as Alzheimer’s disease detection, tumour classification, post-traumatic stress disorder (PTSD) prediction, and brain age estimation \cite{jiang2023self,zhang2016multimodal,zhao2021longitudinal}. However, many existing efforts focus on specific applications, often lacking broader generalisation. Furthermore, although the comparative strengths of contrastive and generative SSL approaches are well studied in natural image domains \cite{chen2020simple,chen2021empirical,he2022masked}, their relative advantages in medical imaging remain underexplored. Additionally, challenges related to interpretability and effective integration of multimodal inputs in SSL models continue to limit their clinical utility in neuroimaging.

To address these challenges, we propose BrainFound, a 3D-aware self-supervised foundation model tailored for brain MRI. It supports both single- and multimodal scans, generalises across varying resolutions, and achieves strong performance in disease detection and image segmentation. We choose DINOv2 \cite{oquab2023dinov2} due to its strong invariance learning through self-distillation, which is particularly advantageous for heterogeneous medical imaging data with variable acquisition protocols. Built by extending the 2D DINOv2 framework to volumetric brain MRI via a slice-wise strategy, BrainFound approximates long-range anatomical context and captures detailed brain structures crucial for accurate analysis.

BrainFound demonstrates strong performance across tasks, effectively handling both diffuse neurodegenerative changes and focal pathologies, such as tumours. By combining general visual priors with domain-adapted SSL, BrainFound establishes a strong benchmark for 3D SSL in brain imaging. We comprehensively evaluate BrainFound across classification, segmentation, cross-dataset generalisation, and few-shot learning tasks spanning neurodegenerative, oncological, and fetal brain imaging.

Our key contributions are:

1.	We designed BrainFound, a 3D SSL model for brain MRI, developed by extending the DINOv2 \cite{oquab2023dinov2} framework to 3D, capturing complex anatomical structures.

2.	We introduce a model architecture that effectively integrates both single- and multimodal MRI inputs (in our current implementation, T1, T2, FLAIR, but adaptable to alternative/more contrasts), thereby enhancing robustness and enabling versatile diagnostic applications.

3.	We demonstrate consistent performance across varying voxel resolutions, including both isotropic (e.g., 1-2 mm³) and anisotropic voxel sizes, which improves generalisability and practical applicability across diverse clinical imaging protocols.

4.	We develop a multitask 3D SSL foundation model capable of supporting a broad spectrum of brain MRI tasks, including multi-disease classification (e.g., multiple dementia types and tumour grading) and multi-region anatomical and tumour segmentation, demonstrating unified performance across both diagnostic and structural analyses.

\section{Material and Methods}

In this study, we develop BrainFound, a versatile self-supervised learning (SSL) foundation model applicable to both single-modality and multimodal brain MRI. Built on large-scale unlabeled data, BrainFound is designed to learn rich, generalisable representations that support a wide spectrum of downstream tasks, including diverse disease detection and detailed anatomical/pathological segmentation. The following sections detail the model architecture, training strategy, and evaluation protocols used to demonstrate its effectiveness across multiple clinical applications.

We chose the pre-trained DINOv2 \cite{oquab2023dinov2} model as the backbone of our self-supervised learning (SSL) approach because of its proven effectiveness in unsupervised representation learning. It involves sequential training on 142 million carefully curated natural images, showcasing significant efficacy across diverse vision tasks. While DINOv2 excels in 2D image tasks, it currently lacks the capability to capture complex spatial relationships in 3D data, such as MRI scans. To address this limitation, we adapted DINOv2 for 3D medical imaging. Subsequently, we extend this approach to multi-contrast brain MRI images from twelve publicly available datasets, encompassing a total of 10,000 volumetric brain MRI images across T1, T2, and FLAIR modalities.

Our proposed model leverages the properties of three MRI modalities/contrasts, T1, T2, and FLAIR, by stacking them as channel-wise inputs, similar to how RGB images are processed in deep learning models. This approach can be easily extended to incorporate additional MRI modalities or contrasts by increasing the number of input channels.

 \begin{figure}[H]
    \centering
    \includegraphics[width=1\textwidth]{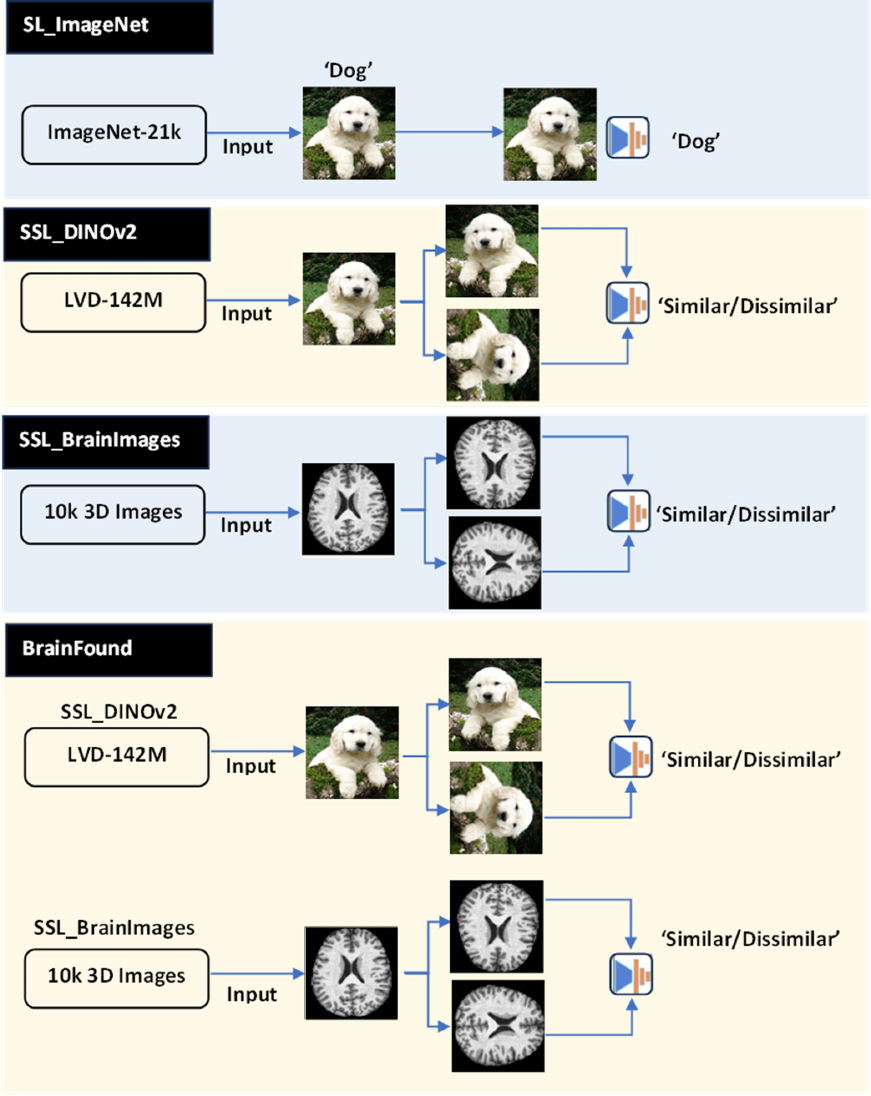}
    \caption{Overview of the training pipeline for BrainFound and baseline models. The baseline models comprised SL\_ImageNet, SSL\_DINOv2, and SSL\_BrainImages. SL\_ImageNet uses supervised learning on ImageNet-21k, which includes 14 million images with categorical labels. SSL\_DINOv2 was pretrained on LVD-142M dataset. SSL\_BrainImages employs self-supervised learning from scratch on brain images. BrainFound, on the other hand, utilises self-supervised learning on brain images, starting from the weights pre-trained on SSL\_DINOv2. *A dog image is used to demonstrate the natural image LVD-142M dataset.
}
    \label{fig:fig1}
\end{figure}

This approach enables the model to extract complementary features from each modality, enhancing representation learning. Additionally, BrainFound is designed to work with both a single modality and a combination of modalities, allowing for a flexible partial modality setting.

\begin{figure}[H]
    \centering
    \includegraphics[width=1\textwidth]{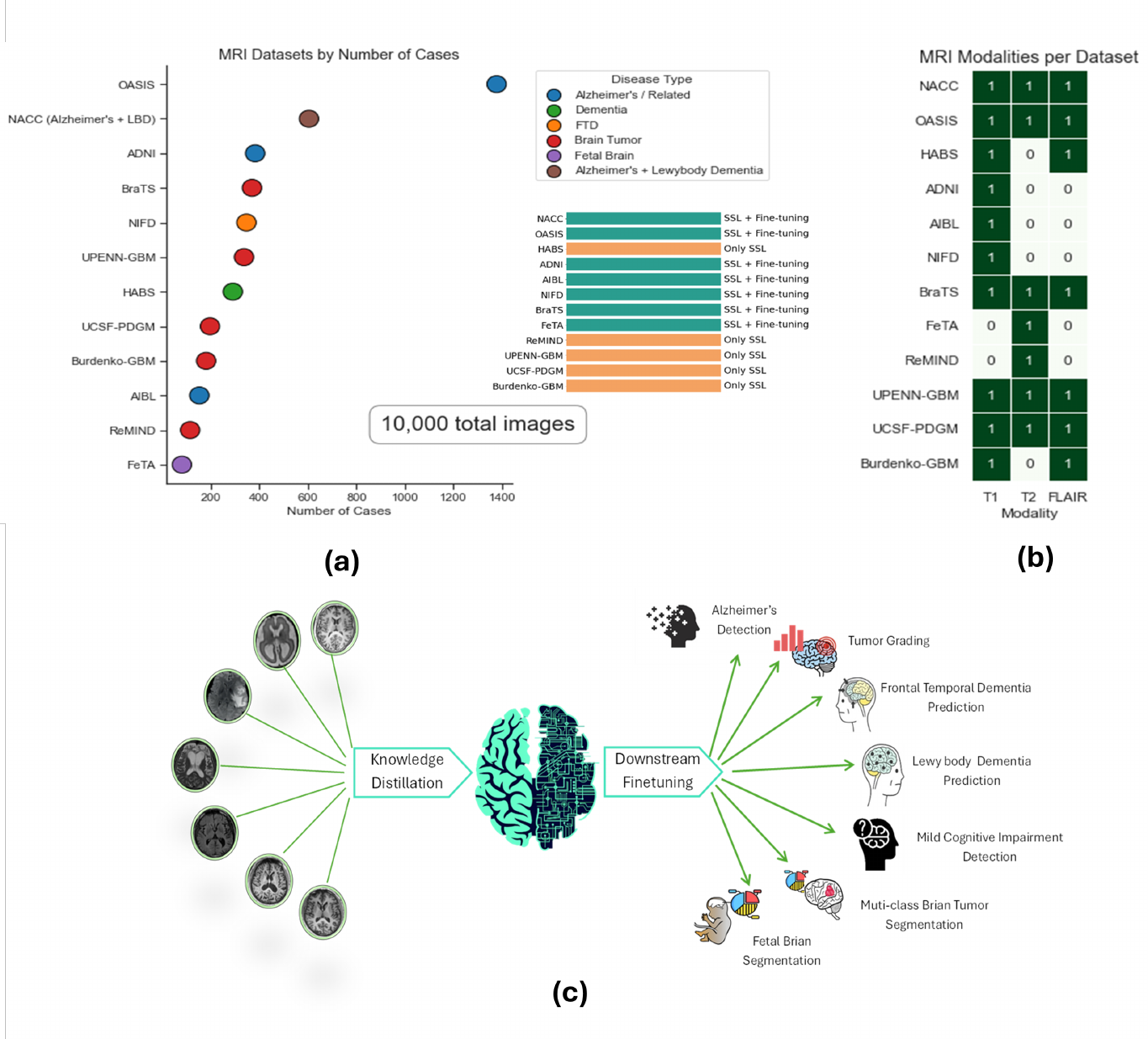}
    \caption{Overview of MRI datasets and modalities, for the SSL and downstream workflow. (a) Number of MRI cases per dataset, color-coded by disease type, including Alzheimer’s disease, frontotemporal dementia (FTD), brain tumours, and fetal brain conditions. Usage of each dataset in the pipeline: whether used for self-supervised learning (SSL) only or for both SSL and fine-tuning. (b) Imaging modality availability (T1, T2, FLAIR) for each dataset, indicating multimodal coverage relevant for pre-training. (c) Diagram of the overall framework: multiple datasets and modalities are used in a knowledge distillation-based SSL pretraining stage, followed by downstream fine-tuning showcasing generalisation across disease detection and image segmentation tasks.
}
    \label{fig:fig2}
\end{figure}

\subsection{Datasets}
For the construction of BrainFound, we curated \~10,000 volumetric brain MRI images of three different modalities, T1, T2, and FLAIR, which were taken from twelve public datasets, including BraTS \cite{baid2021rsna}, ReMIND \cite{juvekar2024remind}, UPENN-GBM \cite{bakas2022university}, UCSF-PDGM \cite{calabrese2022university}, Burdenko-GBM-Progression \cite{zolotova2023burdenko}, HABS \cite{dagley2017harvard}, OASIS \cite{lamontagne2019oasis}, ADNI \cite{petersen2010alzheimer}, NACC \cite{beekly2007national}, AIBL \cite{ellis2009australian}, NIFD \cite{rohrer2013neuroimaging}, and FeTA \cite{payette2021fetal}, as given in Figure \ref{fig:fig1}.

For datasets used in both self-supervised pretraining and downstream fine-tuning, 15\% of the volumes were first separated as a held-out test set to ensure that these images were never seen during pretraining or fine-tuning, preventing any data leakage (see Table \ref{tab:dataset_distribution}. The remaining 85\% of volumes were used for self-supervised learning (SSL) and downstream tasks. For downstream tasks, the remaining 85\% of the data were used for five-fold cross-validation, where in each fold a model was trained on the training split and validated on the corresponding validation split. The details for training/validation and testing splits for each dataset are given in Table \ref{tab:dataset_distribution} of Appendix B.

Only T1-weighted, T2-weighted, and FLAIR images with complete volumetric coverage were included. Images with severe artifacts, motion blur, or incomplete acquisition were excluded. 

Distribution and Demographics: The datasets include subjects across a wide age range, from young adults ($\sim$ 18 years) to older adults ($\sim$ 90 years). Sex distribution varies across cohorts, with some datasets approximately balanced and others showing a slight male predominance, depending on the disease population. Diagnostic classes were balanced whenever possible; for datasets with uneven class distributions, balancing strategies such as random sampling or stratified selection were applied. 

Our framework works for partial scan settings to facilitate all sets of datasets that have a single MRI modality/contrast or hold multimodal datasets, including T1, T2, and FLAIR scans. The details of the scan types, dataset usage, and modality coverage are summarised in Figure \ref{fig:fig2}(a–c). All datasets were processed through standardised preprocessing steps derived from \cite{dorent2021learning}. The steps are: 1) co-registering modalities in the MNI space, and 2) performing skull-stripping on the co-registered imaging modalities.

\subsection{SSL Model Architecture}
DINOv2 (architecture details are introduced in Figure \ref{fig:dinov2} in Appendix A) is built on a Vision Transformer (ViT) backbone, which processes images by dividing them into patches and encoding spatial relationships through self-attention mechanisms. Unlike traditional convolutional neural networks, which focus on local features, Vision Transformers capture global dependencies within an image, making them highly effective for understanding complex structures in brain scans. This architecture is further enhanced by DINO v2’s self-distillation process, where a teacher network generates stable feature representations, and a student network learns from these outputs. Since no labeled data is needed, the model autonomously discovers informative patterns, making it a powerful tool for medical imaging analysis.

\begin{figure}[H]
    \centering
    \includegraphics[width=1\textwidth]{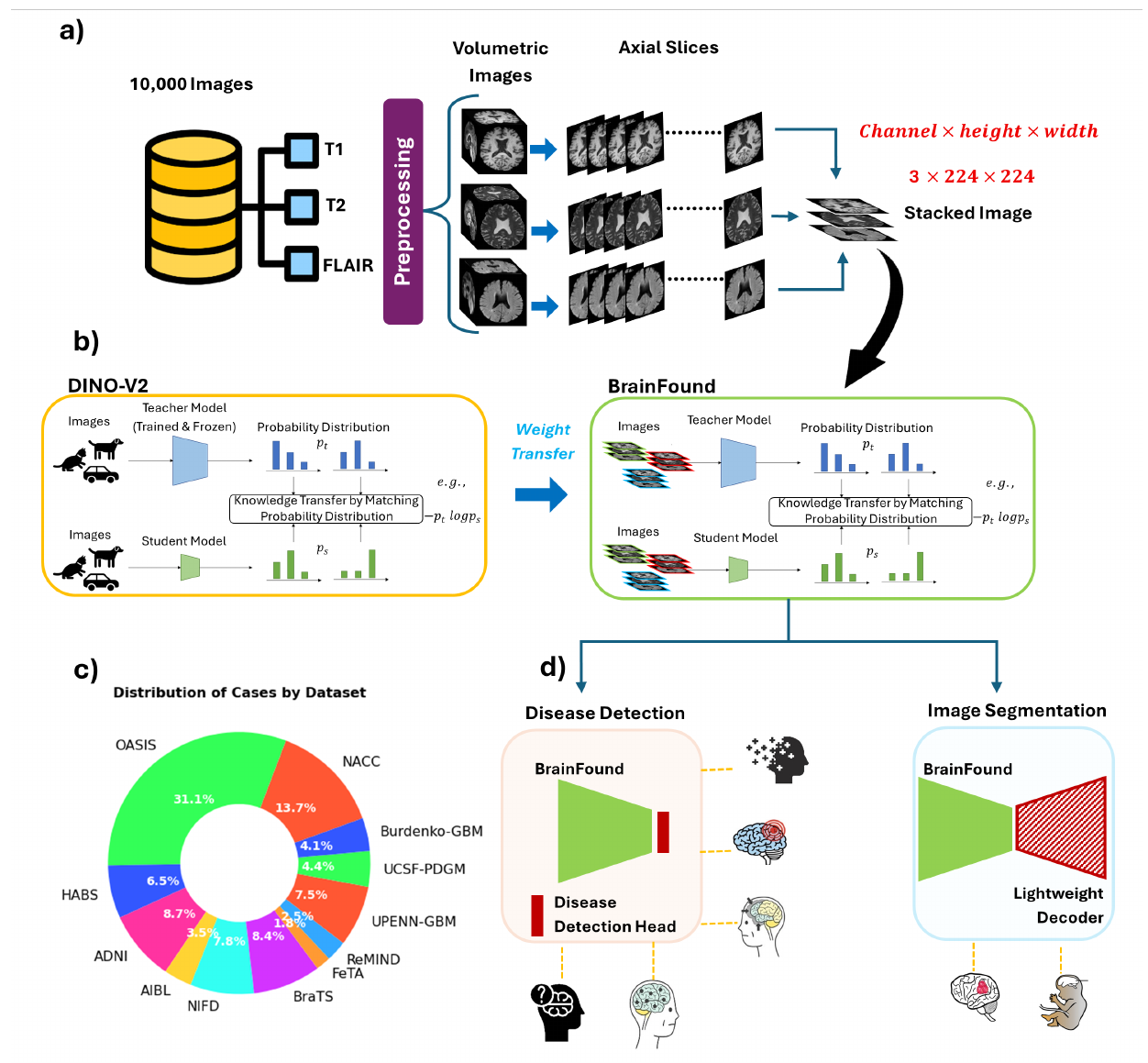}
    \caption{Schematic architecture of BrainFound. (a) Multi-contrast MRI volumes (T1, T2, FLAIR) collected from 12 public datasets ($\sim$10,000 scans) undergo preprocessing and are converted into stacked axial slice representations (3 × 224 × 224). (b) A pretrained DINOv2 teacher–student SSL framework provides initial weights to BrainFound, which is then adapted through domain-specific SSL on stacked MRI slices using probability-distribution matching. (c) Distribution of cases across contributing datasets, illustrating the heterogeneity and scale of the training corpus. (d) Downstream applications: BrainFound serves as a backbone for disease-detection models and as an encoder for lightweight image-segmentation decoders.
}
    \label{fig:framework}
\end{figure}

Our proposed BrainFound model leverages self-supervised learning on volumetric brain MRI by adapting the DINOv2 framework. We employ a ViT-Large backbone and follow the exact training procedure as DINOv2, but applied to MRI data using transfer learning \cite{raghu2019transfusion,yamashita2018convolutional,zhou2023foundation,baharoon2023evaluating}. To handle volumetric and multimodal MRI, we convert 3D volumes into 2D slices while preserving spatial alignment across modalities.  We initialize the ViT-Large backbone with pretrained DINOv2 weights from natural images. No modifications are made to the loss function or optimization strategy of DINOv2; adaptations are limited to data representation and preprocessing to accommodate volumetric MRI. This transfer enables the model to leverage general visual features, which are then adapted to MRI-specific anatomical structures through self-supervised learning. A detailed algorithm of BrainFound pseudocode is given in Figure \ref{fig:pseudocode} in Appendix A.

\begin{figure}[H]
    \centering
    \includegraphics[width=1\textwidth]{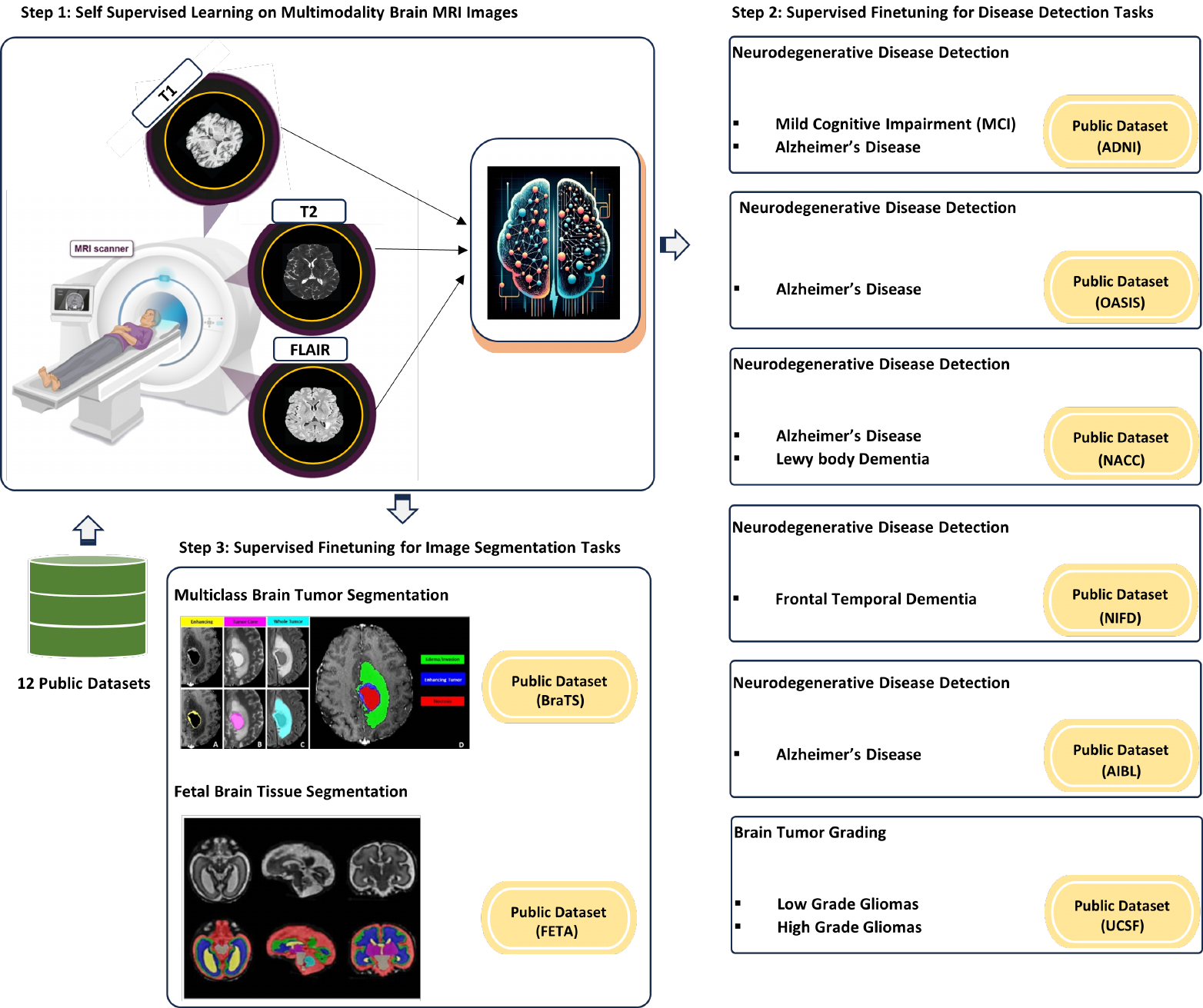}
    \caption{An overview of the development and application of BrainFound. Stage one involves pretraining BrainFound using Self-supervised learning on large-scale multi-scan brain MRI datasets. Stage two adapts the pretrained model to multiple downstream disease detection and image segmentation tasks through supervised fine-tuning, followed by both internal and external evaluation to assess generalisability across datasets.
}
    \label{fig:fig4}
\end{figure}

One of the key advantages of utilising DINOv2 to make BrainFound is its ability to perform multi-scale feature extraction. It learns to capture both fine-grained details and high-level structures in brain imaging data, which is crucial for identifying subtle patterns associated with neurodegenerative diseases. Additionally, by stacking T1, T2, and FLAIR scans as channel-wise inputs like RGB images, the model learns robust, modality-aware feature representations. This design enables BrainFound to generalise across different datasets and varying modality availability, making it highly adaptable to diverse brain research tasks.

Moreover, by training on a large, heterogeneous dataset with varying acquisition parameters, including differences in resolution, scanner vendors, and contrast settings, BrainFound learns robust, modality-invariant representations. This robustness helps the model maintain strong performance despite variations in image quality and contrast characteristics, commonly encountered in clinical and research MRI scans. Moreover, by employing DINOv2 as our self-supervised learning framework, BrainFound achieves effective analysis of brain imaging data while minimizing dependence on extensively labeled datasets.

BrainFound extends the 2D DINOv2 framework to volumetric brain MRI using a sequential slice-wise strategy. Each 3D MRI scan $V \in \mathbb{R}^{H \times W \times D \times C}$ is decomposed into $D$ axial slices. During training, we iterate over the slice depth dimension $d = 1, \ldots, D$, treating each slice as an independent 2D image.

To provide flexibility, BrainFound allows the user to specify the number of slices per volume used for training, enabling variable volumetric depth and efficient handling of heterogeneous MRI datasets. The framework can either evenly sample a fixed number of slices or use all slices from each volume, controlled by the user-configurable parameter \texttt{num\_slices}. For a given volume $V \in \mathbb{R}^{H \times W \times D \times C}$, if $\texttt{num\_slices} \leq D$, the model evenly samples \texttt{num\_slices} slices along the depth dimension; otherwise, all $D$ slices are used.

BrainFound also supports multimodal brain MRI by stacking multiple contrasts as input channels, analogous to RGB channels in natural images. For each subject, corresponding T1-weighted, T2-weighted, and FLAIR slices are stacked along the channel dimension, forming a multichannel 2D input tensor $I \in \mathbb{R}^{H \times W \times m}$. Here, $m$ denotes the number of MRI modalities and corresponds to the number of input channels $C$ for multimodal scans. For single-modality scans, the same modality is duplicated across three channels to satisfy the 3-channel input requirement (see Figure \ref{fig:framework}).

\subsection{Baseline Models}

To evaluate the performance and label efficiency of BrainFound, we compared it against three baseline models: SL\_ImageNet, SSL\_DINOv2, and SSL\_BrainImages. All models share the same architecture and downstream fine-tuning procedures; however, they differ in their pretraining strategies. Notably, BrainFound initializes from DINOv2 weights pretrained via self-supervised learning on diverse natural images from the LVD-142M dataset, rather than being limited to ImageNet.

SL\_ImageNet follows a conventional transfer learning paradigm, with supervised pretraining on ImageNet-21k. SSL\_DINOv2 employs self-supervised learning on approximately 140 million natural images from the LVD-142M dataset~\cite{oquab2023dinov2}. SSL\_BrainImages, in contrast, is trained from scratch using self-supervised learning exclusively on brain MRI data, including T1-weighted, T2-weighted, and FLAIR images, enabling the model to learn from diverse MRI contrasts.

BrainFound leverages DINOv2 pretrained weights as an initialization and further adapts the model through self-supervised learning on brain MRI data. We hypothesise that this strategy benefits from pretraining on natural images, which capture a broad range of textures, shapes, and visual patterns that generalise across domains. By first learning such generic representations and subsequently adapting to domain-specific multimodal brain MRI data (T1, T2, and FLAIR), BrainFound can more effectively support neuroimaging tasks such as disease detection, classification, and segmentation.

The different pretraining strategies are illustrated in Figure \ref{fig:fig1}, while an overview of the development and application of BrainFound is shown in Figure \ref{fig:fig4}.

\subsection{Training Strategy}
We performed self-supervised learning (SSL) by retraining the pretrained DINOv2 model on a dataset of $\sim$10,000 brain MRI volumes (see Figure \ref{fig:fig2}). To adapt the 2D DINOv2 backbone for 3D MRI data, each volumetric scan was processed as a sequence of 2D axial slices, allowing the model to learn spatially aware representations while remaining compatible with the 2D pretraining framework. For all experiments, we used ViT-Large as the backbone.

To capture both global and local brain structures, we applied a multi-scale cropping strategy, generating global crops of 96×96 and local crops of 64×64, with 150 slices per volume. All slices were processed individually through the 2D DINOv2 model with a batch size of 200 slices per iteration, using mixed precision (FP16) for memory efficiency. We employed the full DINOv2 augmentation pipeline, including multi-crop, color jitter, random crop/resize, and Gaussian blur. The average iteration time was approximately 0.5 seconds per batch.

With $\sim$10,000 MRI volumes ($\sim$1.5 million slices total), this configuration resulted in approximately 7,500 iterations per epoch, corresponding to an estimated training time of $\sim$62.5 minutes per epoch. For 100 epochs, total SSL pretraining required roughly 104 hours ($\sim$4.3 days). The model was optimized using AdamW with a learning rate of $1\times10^{-4}$, $\beta_1 = 0.9$, $\beta_2 = 0.95$, and weight decay of 0.05. We followed the DINOv2 self-distillation loss (teacher-student with centering and sharpening), with the teacher network updated via an exponential moving average of the student parameters. A cosine learning rate schedule with linear warmup over the first few epochs was applied to improve training stability. This setup enabled efficient learning of slice-wise 2D self-supervised representations while preserving both global and local structural information.

For downstream tasks, we applied data augmentations including flipping, scaling, brightness and contrast adjustments, and advanced transformations such as RandGaussianRotate, RandZoomd, RandShiftIntensity, and RandGaussianNoise~\cite{MONAI2022}. The downstream loss function combined cross-entropy for disease detection and a combination of Dice loss and cross-entropy for segmentation, optimizing performance across both tasks. The AdamW optimizer was used with a learning rate of $1\times10^{-4}$. To prevent overfitting and accelerate training, early stopping was employed based on validation metrics, with training typically concluding around 80 epochs. Downstream training took less than 10 hours with a batch size of 20 on an NVIDIA A6000 GPU with 48,GB of memory.

To evaluate model performance and generalisability, we implemented a five-fold cross-validation scheme on the training data, while maintaining a strictly held-out test set for final evaluation. From each dataset, 15\% of the volumes were held out as a final test set and were never used during self-supervised pretraining, fine-tuning, or validation. Importantly, all reported test results are obtained exclusively on these strictly held-out volumes.

The remaining 85\% of the data was used for model development and cross-validation. This training data was partitioned into five stratified folds of equal size. In each fold, four folds (approximately 70\% of the total data) were used for training, and one fold (approximately 15\%) was used for validation to support model selection and early stopping. This process was repeated across all five folds, resulting in five distinct training–validation splits.

For each fold, the model was trained on the training split and validated on the corresponding validation split after each epoch. The model checkpoint with the best validation performance was saved per fold. After completing all folds, five trained models were obtained—one per fold. Each saved model was then evaluated on the same held-out test set (15\% of the data). Performance metrics such as AUROC and Dice score were computed for each fold’s model, and the mean and standard deviation across folds were reported. Statistical significance was assessed using a two-sided t-test on the performance metrics across the five folds \cite{zhou2023foundation}.

\subsection{Downstream Tasks and Evaluation Scenarios}
BrainFound is evaluated on a diverse set of downstream tasks designed to test classification, segmentation, generalisation, and label efficiency. These include (i) multi-cohort neurodegenerative disease detection, (ii) brain tumour grading, (iii) external cross-dataset evaluation, (iv) few-shot learning, and (v) anatomical and pathological brain segmentation. After self-supervised pretraining on these multi-sequence/multimodality brain MRI images, we evaluated the performance and generalisability of BrainFound in adapting to diverse brain imaging tasks using a partial modality setting (single or combined modality). We selected publicly available datasets for the multiple neurodegenerative disease detection tasks, including AD (NACC, ADNI, OASIS, and AIBL datasets), MCI (ADNI dataset), LBD (NACC dataset), and FTD (NIFD dataset) detection. Other than the neurodegenerative tasks, we also tested BrainFound on a brain cancer imaging dataset for low and high-grade tumour grading (UCSF dataset). Moreover, we trained and validated BrainFound for image segmentation tasks, including brain tumour (BraTS dataset) subtype lesion segmentation as well as fetal brain anatomical sub-regions segmentation (FeTA dataset). The details of these datasets are given in Figure \ref{fig:fig2}.

\paragraph{Task 1: Disease Detection}

To comprehensively evaluate BrainFound’s clinical utility, we benchmarked its performance across a range of disease detection tasks involving both neurodegenerative and oncological conditions. For neurodegenerative disorders, five publicly available datasets (AD (NACC, ADNI, OASIS, and AIBL), MCI (ADNI), LBD (NACC), and FTD (NIFD)) were used to assess classification performance across Alzheimer’s disease, frontotemporal dementia, mild cognitive impairment, and Lewy body dementia. To test generalisability beyond neurodegeneration, we applied BrainFound to an interdisciplinary dataset focused on brain tumour grading. This dataset included adult patients with histopathologically confirmed grade II–IV diffuse gliomas, who underwent preoperative MRI and tumour genotyping at a single academic center between 2015 and 2021 \cite{calabrese2022university}. Across all disease detection tasks, each model’s discriminative power was assessed with AUROC (Area Under the ROC Curve) as the primary metric to evaluate classification performance. We also used a statistical two-sided t-test to compare AUROC values from five-fold cross-validation across models, revealing significant differences (p < 0.05). A p-value below the conventional threshold of 0.05 allows us to confidently reject the null hypothesis, supporting the effectiveness of the BrainFound model over other training approaches in achieving superior performance.

\paragraph{Task 2: Cross-dataset External Evaluation}
	
To evaluate the robustness and generalisability of BrainFound, we performed cross-dataset evaluations using independent external datasets for the AD disease detection task that were not part of the training process. Unlike internal validation, which uses data splits from the same datasets, this evaluation tests the model’s ability to maintain high classification accuracy on completely unseen data acquired under different imaging protocols and conditions. This setup better reflects real-world clinical scenarios, where variability in scanner types, acquisition parameters, and patient populations can impact model performance.

\paragraph{Task 3: Evaluation on Few-Shot Learning}
	 
To assess the models’ ability to generalise with limited labeled data, we conducted few-shot learning experiments. For each downstream task, we trained the models using varying fractions of the available labeled training data, ranging from 10\% to 100\% (e.g., 10\%, 20\%, 50\%, 100\%). The performance was then evaluated on the full test set to measure how well the models learn from scarce annotations. This approach reflects practical clinical scenarios where labeled data is often limited. Detailed evaluation metrics and results are reported in the ‘Experiments and Results’ section.

\paragraph{Task 4: Brain MRI Segmentation}

After completing disease detection tasks, we further enhanced the capabilities of the BrainFound model to address brain image segmentation tasks. For the image segmentation task, we evaluated the model’s performance by measuring the Dice and HD95 scores for BrainFound alongside other baseline and state-of-the-art models.

\section{Results Across Classification, Generalisation, Few-Shot, and Segmentation Tasks}
We evaluate BrainFound on two core neuroimaging tasks: disease classification and brain structure/pathology segmentation. The model’s performance is compared against both supervised and self-supervised baselines across multiple public datasets and modalities (T1, T2, FLAIR).

For disease detection, we benchmark BrainFound against SL\_ImageNet (supervised ImageNet pretraining), SSL\_DINOv2 (self-supervised on LVD-142M)~\cite{oquab2023dinov2}, and SSL\_BrainImages (self-supervised on brain MRI data). Additional comparisons include SimCLR~\cite{chen2020simple}, SwAV~\cite{karnyoto2024swav}, MoCo-v3~\cite{chen2021empirical}, and Masked Autoencoder~\cite{he2022masked} to assess the benefits of the DINOv2-based self-distillation approach.

For segmentation, we evaluate BrainFound against strong baselines such as U-Net and state-of-the-art models, including MedSAM~\cite{ma2024segment}, Hi-End-MAE~\cite{tang2025hi}, SSL\_DINOv2, and SSL\_BrainImages. Evaluations were conducted across multiple datasets using Dice and HD95 as standard metrics. Detailed results and comparisons for both tasks are presented below.

\subsection{Disease Detection Task}
This section presents the evaluation of BrainFound in detecting neurological disorders, specifically neurodegenerative diseases, and brain tumour grading.

\paragraph{Task 1: Alzheimer’s Disease Detection}

\begin{figure}[H]
    \centering
    \includegraphics[width=1\textwidth]{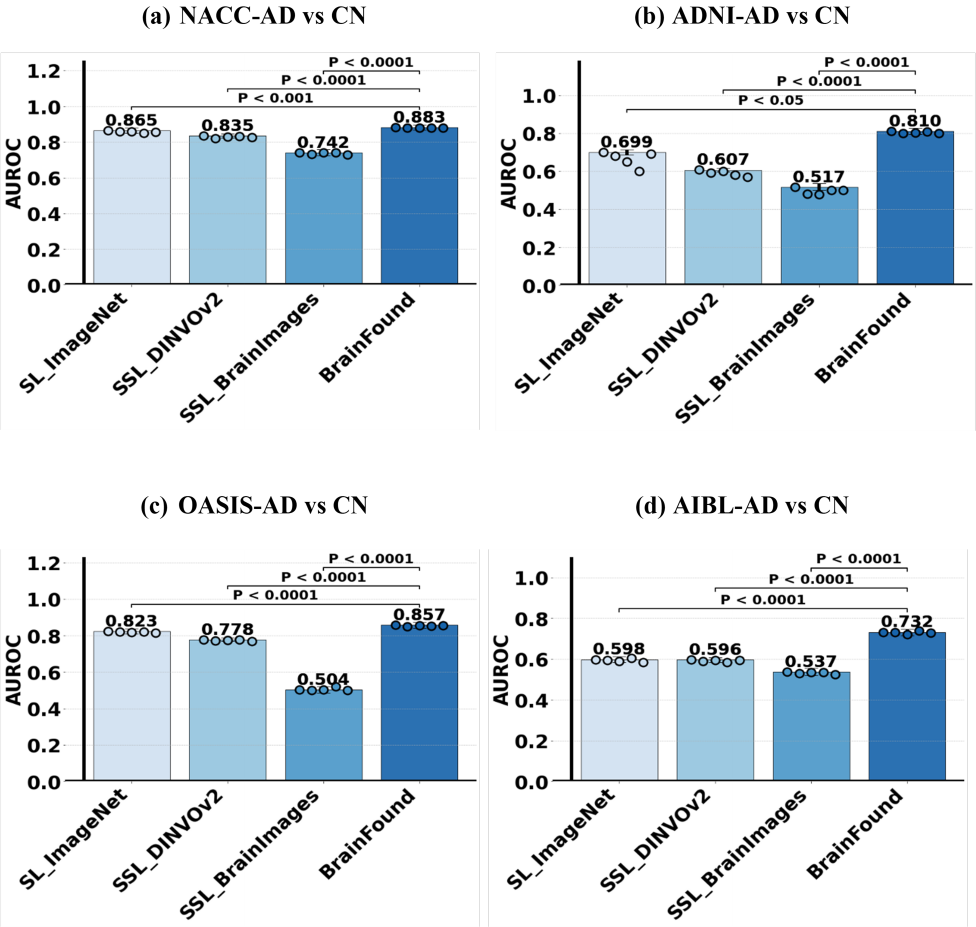}
    \caption{AUROC comparison across pretraining strategies for Alzheimer's disease detection. Models were initialized with one of four pretraining methods: SL\_ImageNet, SSL\_DINOv2, SSL\_BrainImages, or the proposed BrainFound. Each task was evaluated using 5-fold cross-validation. The mean AUROC across folds is shown as the height of each bar. Individual fold results are indicated by the five dots on each bar, illustrating variability across folds. BrainFound consistently outperforms the baseline methods, with statistically significant improvements (P $<$ 0.05, two-sided t-test~\cite{zhou2023foundation}) indicated above the bars.
}
    \label{fig:fig5a}
\end{figure}

To assess the clinical robustness and generalisability of BrainFound, we evaluated its performance in Alzheimer’s disease (AD) detection across four independent cohorts: NACC, ADNI, OASIS, and AIBL. In each case, the model was tasked with distinguishing individuals diagnosed with AD from cognitively normal (CN) controls, using internal validation protocols wherein training and testing sets were drawn from the same dataset. BrainFound consistently outperformed competing models across all four datasets, achieving the highest area under the receiver operating characteristic curve (AUROC) in each task (Figure \ref{fig:fig5a}). On the NACC dataset, BrainFound achieved an AUROC of 0.883, markedly surpassing SL\_ImageNet (0.865), SSL\_DINOv2 (0.835), and SSL\_BrainImages (0.742). The high accuracy in this task highlights BrainFound’s ability to capture disease-specific structural changes in the brain, likely enabled by its dual-stage training: general representation learning via DINOv2 pretraining on natural images, followed by domain adaptation through self-supervised learning on multimodal brain MRI scans (T1, T2, FLAIR).

Similarly, in the ADNI dataset, BrainFound achieved an AUROC of 0.810, significantly outperforming SL\_ImageNet (0.699), SSL\_DINOv2 (0.607), and SSL\_BrainImages (0.517). Performance gains persisted across the OASIS dataset, where BrainFound reached an AUROC of 0.857, compared to SL\_ImageNet (0.823), SSL\_DINOv2 (0.778), and SSL\_BrainImages (0.504). Finally, in the AIBL cohort, BrainFound again led with an AUROC of 0.732, while SL\_ImageNet (0.598), SSL\_DINOv2 (0.596), and SSL\_BrainImages (0.537) lagged.

\paragraph{Task 2: Lewy Body Dementia Detection}

\begin{figure}[H]
    \centering
    \includegraphics[width=1\textwidth]{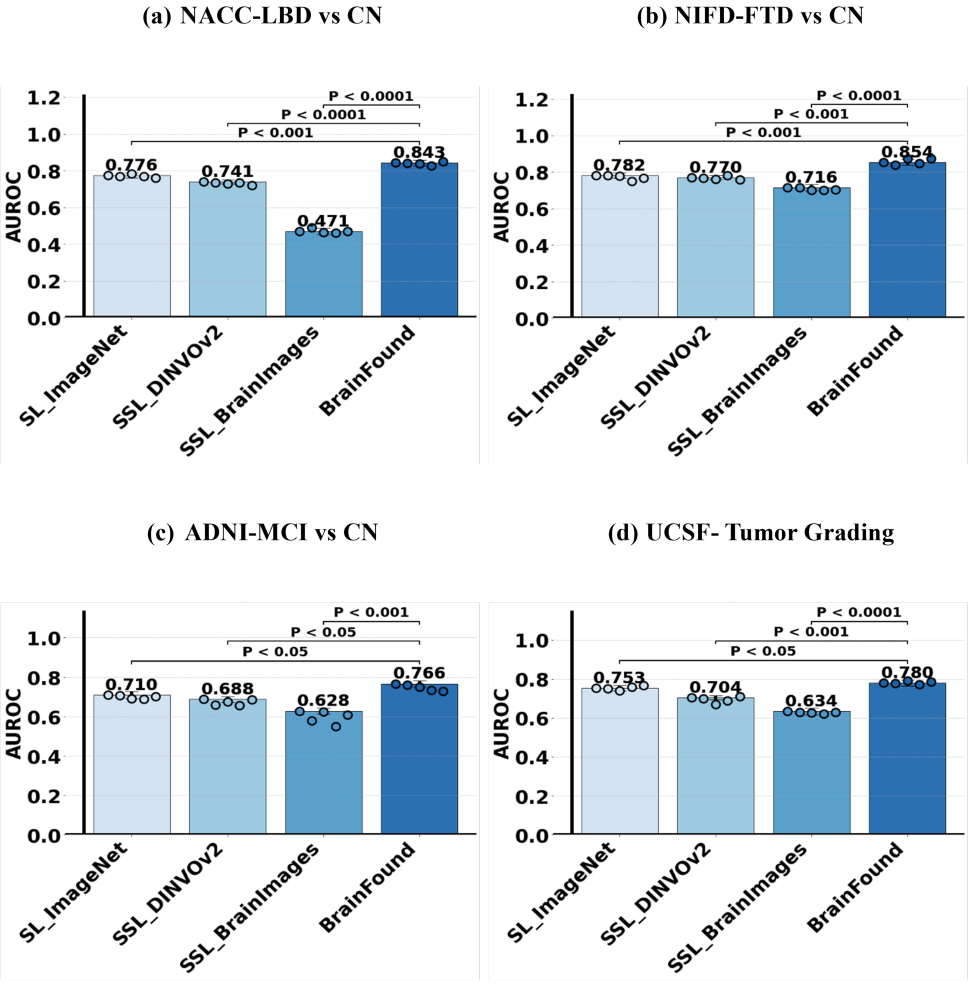}
    \caption{Comparison of AUROC performance across different pretraining strategies on multiple disease detection tasks: Lewy body dementia (a), frontotemporal dementia (b), mild cognitive impairment (c), and tumor grading (d). Each panel shows AUROC scores for models initialized with one of four pretraining methods: SL\_ImageNet, SSL\_DINOv2, SSL\_BrainImages, or the proposed BrainFound approach. BrainFound consistently outperforms the baseline methods, with statistically significant improvements (P $<$ 0.05, two-sided t-test~\cite{zhou2023foundation}) indicated above the bars. Dots above each bar represent AUROC values from five cross-validation folds, illustrating performance robustness and variability.
}
    \label{fig:fig5b}
\end{figure}

The classification task on the NACC dataset was expanded to differentiate Lewy Body Dementia (LBD) patients from cognitively normal controls (Figure \ref{fig:fig5b}). Here, BrainFound exhibited a strong performance, achieving an AUROC of 0.843, indicating its capability to generalise beyond Alzheimer’s disease to other neurodegenerative conditions. SSL\_DINOv2 and SL\_ImageNet also performed well, with AUROC scores of 0.741 and 0.776, while SSL\_BrainImages showed lower performance, with AUROC of 0.471, respectively.

\paragraph{Task 3: Frontotemporal Dementia Detection}

The evaluation was conducted on the NIFD dataset, which involves distinguishing Frontotemporal Dementia (FTD) patients from cognitively normal individuals. In this very challenging classification task, BrainFound continued to perform well, achieving an AUROC of 0.854, as shown in Figure \ref{fig:fig5b}. SL\_ImageNet followed closely with an AUROC of 0.782, while SSL\_BrainImages and SSL\_DINOv2 achieved AUROCs of 0.716 and 0.770, respectively.

\paragraph{Task 4: Mild Cognitive Impairment Detection}

We have compared the AUROC scores of the variant models for Mild Cognitive Impairment (MCI) from cognitively normal controls by using the ADNI cohort. From Figure \ref{fig:fig5b}, it is seen that the proposed BrainFound achieved the best scores (0.766) in detecting MCI, which underscores BrainFound’s early disease detection capabilities. BrainFound’s superior performance highlights its adaptability to different neurodegenerative diseases and its capacity to manage more challenging disease presentations.

\paragraph{Task 5: Brain Tumour Grading (Oncological Disease Detection)}

In the UCSF-Tumour Grading task, we tested the generalisability of our proposed model on detecting the complex tumour grading task beyond neurodegenerative disease detection. As shown in Figure \ref{fig:fig5b}, BrainFound again demonstrated the highest AUROC, further supporting its superior performance across different types of medical image classification and disease detection tasks. The AUROC for BrainFound in this context was 0.780, reinforcing its ability to differentiate between tumour grades effectively.

The SL\_ImageNet and SSL\_DINOv2 models performed reasonably well with an AUROC of 0.753 and 0.704. SSL\_BrainImages showed the lowest performance, with an AUROC of 0.634. The relatively low AUROC of SSL\_BrainImages can be attributed to training self-supervised learning from scratch on domain-specific brain MRI data, which limits feature richness. BrainFound, by contrast, starts from pretrained DINOv2 weights and further adapts to the brain MRI domain using SSL, enabling it to capture more informative and transferable representations.

BrainFound consistently outperformed baseline models across these heterogeneous cohorts, demonstrating robust sensitivity to disease-specific anatomical patterns. These results establish BrainFound as a versatile foundation model capable of detecting diverse brain pathologies from MRI, with high performance and broad applicability across clinical domains.

\subsection{Cross-dataset External Evaluation}

\begin{figure}[H]
    \centering
    \includegraphics[width=1\textwidth]{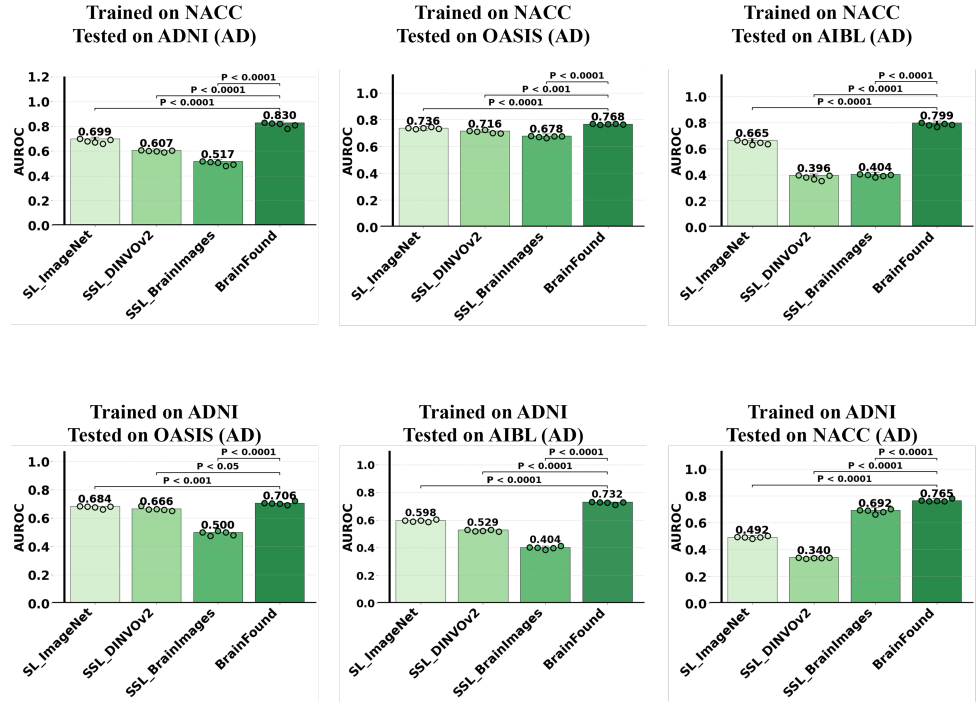}
    \caption{Comparison of AUROC performance across different pretraining strategies under an external evaluation setting, where models are pretrained on one dataset and evaluated on a distinct, unseen dataset. Each panel shows AUROC scores for models initialized with four pretraining methods: SL\_ImageNet, SSL\_DINOv2, SSL\_BrainImages, and the proposed BrainFound approach. BrainFound consistently achieves superior performance across all tasks, significantly outperforming baseline methods (P $<$ 0.05, two-sided t-test), with significance indicated above the bars. Dots above each bar represent AUROC values from individual folds in a five-fold evaluation, illustrating robustness and variability across splits.
}
    \label{fig:fig6}
\end{figure}

To assess external validity, we conducted six cross-cohort evaluations where models trained on one dataset were tested on independent cohorts for Alzheimer’s disease (AD) classification. When trained on the NACC dataset, BrainFound achieved the highest AUROC in all target cohorts: 0.830 on ADNI, 0.880 on OASIS, and 0.799 on AIBL, outperforming all baselines (Figure \ref{fig:fig6}).

Similarly, when trained on ADNI, BrainFound maintained top performance, achieving AUROC scores of 0.765 on NACC, 0.706 on OASIS, and 0.732 on AIBL. In contrast, models pretrained on natural images (SL\_ImageNet, SSL\_DINOv2) or only domain-specific brain MRI datasets (SSL\_BrainImages) consistently underperformed across these cross-domain tasks.

\subsection{State of the Art Model Comparison on Disease Detection Task}
In Figure \ref{fig:fig7} (a) and (b). We compare the AUROC performance of models pretrained using different self-supervised learning (SSL) strategies, including SwAV, SimCLR, MoCo-v3, MAE, and DINOv2, for Alzheimer’s disease detection on the NACC dataset and for tumour grading on the UCSF dataset from multimodality (T1, T2, and FLAIR) MRI imaging data.

The results in Figure \ref{fig:fig7}(a) show that DINOv2 achieves the highest AUROC for Alzheimer’s disease detection on the NACC dataset, outperforming all other self-supervised learning methods. MAE ranks second, demonstrating competitive performance but remaining inferior to DINOv2. SwAV, SimCLR, and MoCo-v3 show comparable but lower performance. These results highlight the effectiveness of DINOv2’s student–teacher self-distillation framework for learning discriminative representations from multimodal brain MRI. In the tumour grading task (Figure \ref{fig:fig7}b), SimCLR slightly outperforms MoCo-v3, while DINOv2 and MAE remain among the top-performing methods.

\subsection{Evaluation on Few-Shot Learning} 
\begin{figure}[H]
    \centering
    \includegraphics[width=1\textwidth]{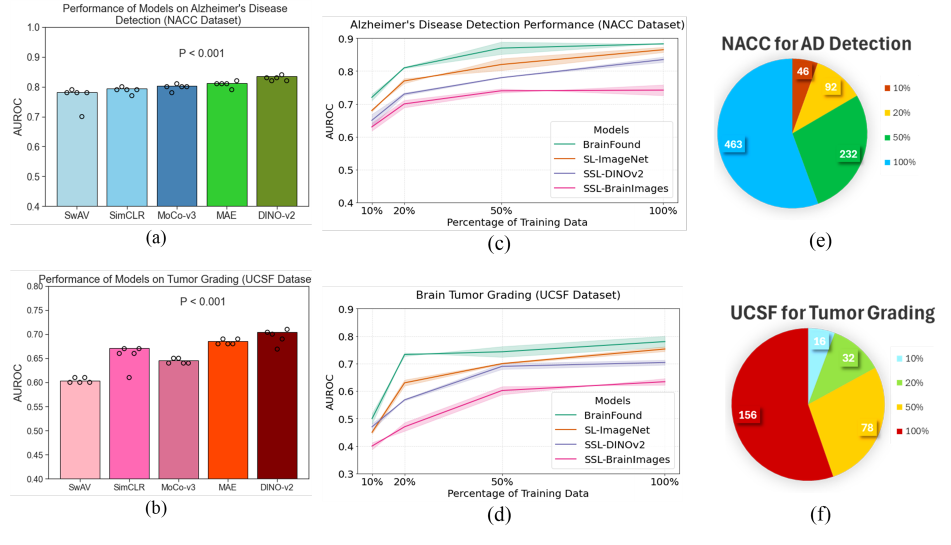}
    \caption{(a) Performance comparison of state-of-the-art foundation models with DINOv2 for Alzheimer’s disease detection on the NACC dataset. (b) Performance comparison of state-of-the-art foundation models with DINOv2 for tumour grading on the UCSF dataset. (c)(d) Few-shot learning comparisons of the baseline models with BrainFound for Alzheimer’s disease detection and tumour grading. (e)(f) Data distribution for few-shot learning.}
    \label{fig:fig7}
\end{figure}

We evaluate the few-shot learning capabilities of different models by measuring their performance across varying fractions of labeled training data (Figure \ref{fig:fig7} (c-d)). This simulates settings where only limited supervision is available, a hallmark of real-world clinical deployment. Our model, BrainFound, demonstrates superior performance even under severe data scarcity: with only 10\% of labeled data, it outperforms the most competitive baseline models. When training data increases to 50\%, BrainFound maintains performance on par with or better than all other models, highlighting its robustness in both few-shot and low-resource regimes.

To better quantify the data required for model convergence, Figure \ref{fig:fig7} (e-f) visualizes the percentage and absolute count of labeled samples used at each stage. The dashed grey lines indicate the relative data usage needed by each model to reach comparable performance. Performance is reported as AUROC, with 95\% confidence intervals visualized as colored bands around the mean AUROC estimates. This behavior indicates that BrainFound learns task-agnostic anatomical priors during self-supervised pretraining, thereby reducing reliance on task-specific labeled data.

\subsection{BRAIN MR IMAGE SEGMENTATION}

\begin{figure}[H]
    \centering
    \includegraphics[width=1\textwidth]{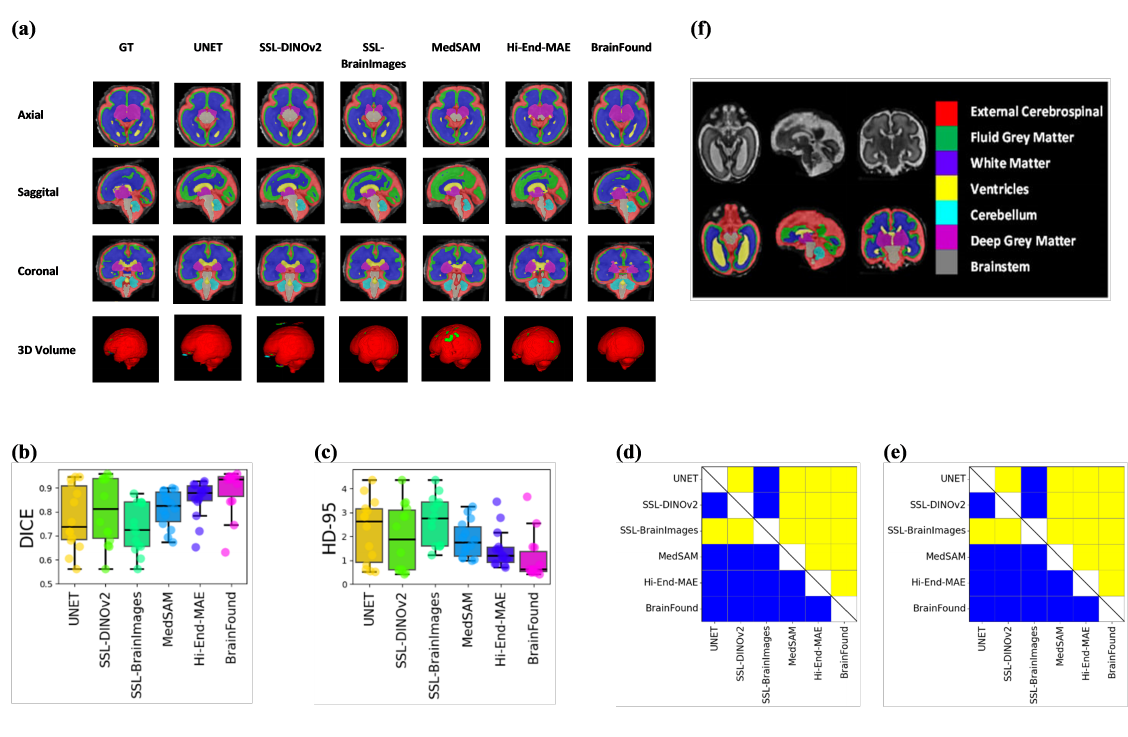}
    \caption{Fetal brain tissue segmentation results from the FeTA dataset. (a) shows the visualisation of different views of the brain (top-down) for ground-truth (GT) and BrainFound predicted masks, along with the prediction of other baseline models, with colored regions representing different brain structures. (b)(c) Boxplots of the Dice coefficient and HD 95 comparing BrainFound with other baseline segmentation methods for the fetal brain tissue segmentation dataset. (d)(e) Pairwise comparison matrix of model performance for Dice (higher is better) and HD95 (lower is better). Blue indicates the model performs better than the compared model (high rank), yellow indicates worse performance (low rank), and white marks the diagonal (self-comparison). (f) shows the class-mapping function for the axial, sagittal, and coronal slices.}
    \label{fig:fig8}
\end{figure}
Building on its strengths in disease detection, we extended BrainFound to two segmentation benchmarks: fetal brain tissue segmentation (FeTA 2021) and brain tumour subtype segmentation (BraTS 2020). These tasks were approached using single-sequence and multi-sequence MRI to evaluate the model’s adaptability across clinical imaging scenarios.

\paragraph{Task 1: Fetal Brain Tissue Segmentation (FeTA 2021)}

Accurate segmentation of fetal brain tissues is vital for assessing development and detecting abnormalities. BrainFound was trained on 80 T2-weighted fetal MRI scans labeled into seven types \cite{payette2021fetal}. Across axial, sagittal, and coronal views, BrainFound outperformed all baseline and state-of-the-art models, producing anatomically faithful segmentations with clear structural boundaries (Figure \ref{fig:fig8}(a)).

Quantitatively, BrainFound achieved the highest median Dice score and lowest HD95, indicating superior volumetric accuracy and boundary precision (Figure \ref{fig:fig8} (b-c)). Statistical comparisons (Figure \ref{fig:fig8} (d-e)) confirmed significant performance gains over all baselines, including Hi-End-MAE and MedSAM, underscoring BrainFound’s reliability and generalisability in low-contrast fetal imaging.

\begin{figure}[H]
    \centering
    \includegraphics[width=1\textwidth]{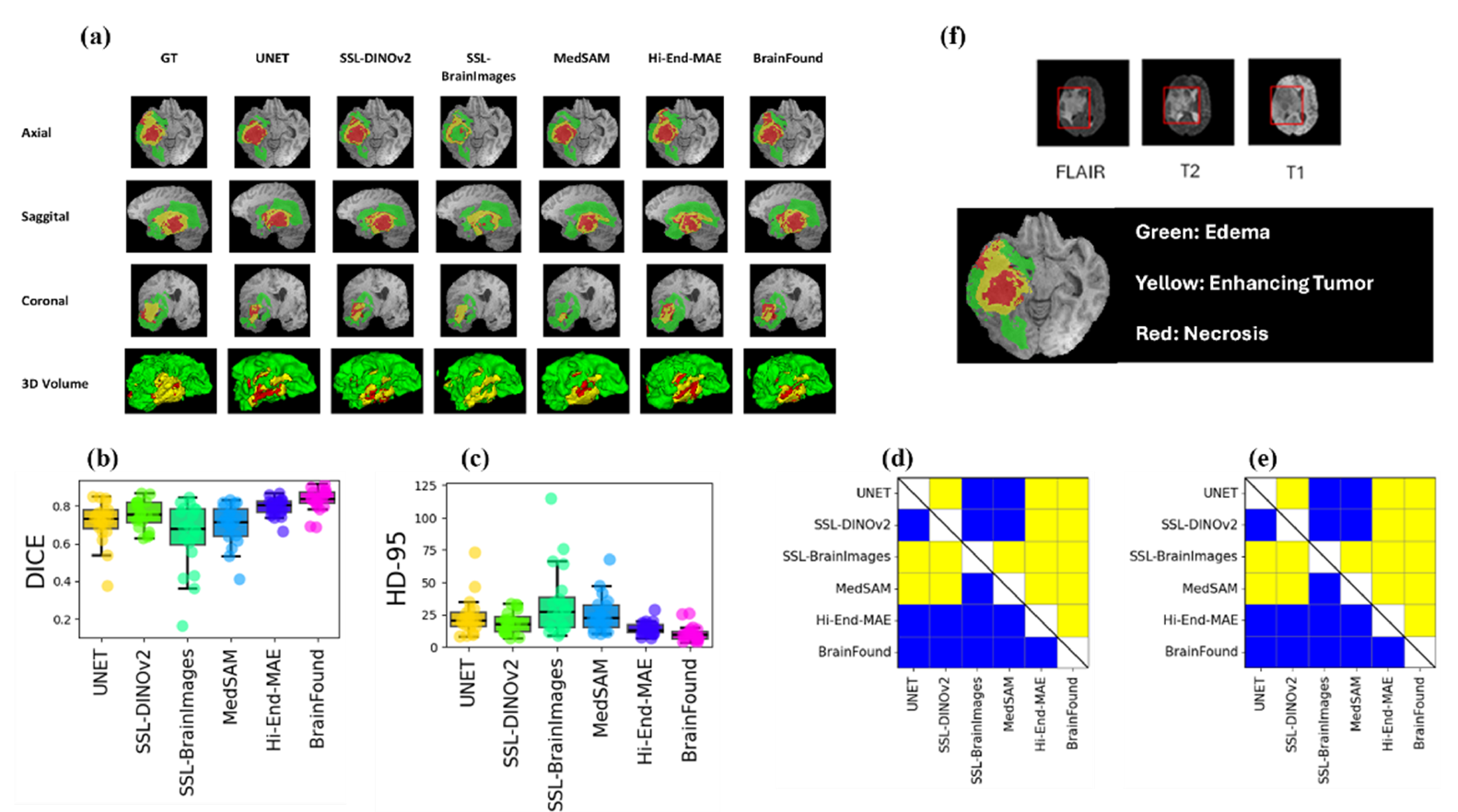}
    \caption{(a) Brain tumour subtype segmentation results from the BraTS dataset. Image (a) shows the visualisations of the different views of the brain (top-down) for ground-truth (GT) and BrainFound predicted masks, along with the prediction of other baseline models, with colored regions representing different brain tumour labels. (b)(c) Boxplots of the Dice coefficient and HD 95 comparing BrainFound with other baseline segmentation methods for the brain tumour subtypes segmentation dataset. (d)(e) Pairwise comparison matrix of model performance for Dice (higher is better) and HD95 (lower is better). Blue indicates the model performs better than the compared model (high rank), yellow indicates worse performance (low rank), and white marks the diagonal (self-comparison). (f) shows T1, T2, T1CE, and FLAIR MRI sequences for a patient with brain tumours are highlighted in red boxes.}
    \label{fig:fig9}
\end{figure}

\paragraph{Task 2: Brain Tumour Subtype Segmentation (BraTS 2020)}

For glioma segmentation, BrainFound was evaluated on the BraTS dataset \cite{oquab2023dinov2} across three tumour subregions: enhancing tumour, edema, and necrotic core. Leveraging multi-contrast MRI (T1, T2, FLAIR), the model achieved accurate, detailed segmentations across 2D and 3D views (Figure \ref{fig:fig9} (a)). BrainFound consistently outperformed baseline and state-of-the-art models in Dice and HD95 metrics, with minimal variance (Figure \ref{fig:fig9} (b-c)). Pairwise statistical tests (Figure \ref{fig:fig9} (d-e)) confirmed the significance of these improvements. Hi-End-MAE \cite{tang2025hi} showed competitive results but did not statistically surpass BrainFound.

\section{Discussion}
This work presents BrainFound, a versatile self-supervised foundation model for volumetric brain MRI that supports multiple downstream tasks, including disease detection and image segmentation across diverse neurological conditions. BrainFound leverages a 3D self-supervised learning (SSL) framework that processes volumetric MRI as sequences of 2D slices, capturing both spatial and anatomical context critical for brain-related tasks. Building on DINOv2, BrainFound extends self-distillation learning to the 3D domain, combining natural image pretraining with brain MRI-specific pretraining to learn both generalizable and domain-specific representations. This slice-wise approach allows the model to retain DINOv2’s powerful 2D feature representations while efficiently modeling 3D anatomical structures, striking a practical balance between performance and computational efficiency.

Early advances in medical imaging relied on supervised pretraining on natural images, with models such as ResNet, DenseNet, and Vision Transformer (ViT) pretrained on ImageNet providing effective initialization for radiology and MRI tasks. Despite substantial differences between natural and medical images, low- and mid-level visual features edges, contours, and textures were shown to transfer effectively, improving convergence and performance in data-scarce scenarios \cite{raghu2019transfusion,yamashita2018convolutional}. More recently, SSL approaches such as SimCLR, BYOL, MAE, and DINOv2 have demonstrated that large unlabeled datasets can yield robust, generalizable representations. In the medical domain, SSL models such as RETFound \cite{zhou2023foundation}, 3D-Heart\_Seg
\cite{qayyum2025foundation} and DINOv2-based studies \cite{baharoon2023evaluating} have shown strong few-shot transfer capabilities across diagnostic tasks, highlighting the scalability of SSL for data-limited settings.

BrainFound builds directly on this paradigm, adapting DINOv2 from 2D natural images to volumetric brain MRI via a slice-wise strategy. Leveraging DINOv2 pretrained representations as initialization, BrainFound is further self-supervised on ~10,000 unlabeled MRI volumes. This hybrid pretraining allows the model to capture general visual features while tuning to domain-specific anatomical structures, supporting strong few-shot and cross-dataset generalization without requiring labeled data. The model accommodates single- and multimodal MRI sequences (T1, T2, FLAIR, contrast-enhanced), enabling integration of complementary anatomical information across modalities, which enhances diagnostic performance.

We benchmarked BrainFound against three baselines for disease detection: (1) SL\_ImageNet, a supervised model pretrained on natural images; (2) SSL\_DINOv2, an off-the-shelf self-supervised model trained on natural images; and (3) SSL\_BrainImages, trained from scratch using MRI self-supervision. While these baselines showed competitive results, none matched BrainFound’s consistency or overall accuracy across tasks, demonstrating the importance of combining natural-image priors with domain-specific pretraining. Comparisons with other SSL methods (SimCLR, MAE, MoCo v3, SwAV, Hi-End-MAE) indicate that DINOv2’s student-teacher self-distillation provides more stable and discriminative representations, particularly beneficial for 3D neuroimaging data.

In segmentation tasks, BrainFound consistently outperformed baseline models (UNet, SSL\_DINOv2, SSL\_BrainImages) and advanced state-of-the-art methods, including MedSAM and Hi-End-MAE, achieving superior Dice and HD95 metrics across single and multi-sequence MRI data. These results underscore BrainFound’s versatility and reliability across diverse neuroimaging scenarios.

Despite its strengths, BrainFound has limitations. Processing volumetric MRI as ordered 2D slices allows efficient use of DINOv2’s 2D feature extractor but does not capture full volumetric context. Additionally, some overlap existed between pretraining and downstream datasets due to limited data availability, which may have slightly inflated development-phase performance but did not affect evaluation on unseen test sets. Future work includes developing fully 3D SSL architectures and expanding pretraining datasets from diverse sources while ensuring strict separation between pretraining and downstream data. BrainFound can also be extended to additional 3D imaging modalities (e.g., T2*, SWI, diffusion maps) and requires broader clinical validation across populations, scanners, and acquisition protocols.

In summary, BrainFound provides a 3D self-supervised learning framework that effectively leverages both natural-image knowledge and domain-specific MRI features. Its strong performance across disease detection and segmentation tasks, adaptability to multimodal inputs, and computational efficiency position it as a robust foundation model for neuroimaging research and future clinical applications.

\section{Conclusion}
BrainFound introduces a self-supervised learning paradigm, marking a significant step toward a generalised foundation model for brain MRI. It effectively captures rich spatial and anatomical features from volumetric MRI data, supporting both single- and multi-sequence inputs. By leveraging a DINOv2-based knowledge-distillation framework and dual-domain pretraining on natural images alongside domain-specific brain scans, BrainFound achieves robust and generalisable performance in a diverse range of disease detection/classification and segmentation tasks. Current limitations include the processing of 2D slices to approximate 3D volumetric inputs, rather than operating on fully volumetric data, and the exclusion of functional or metabolic imaging modalities. Nonetheless, the model’s adaptability to diverse imaging protocols and strong comparative performance underscore its potential as a versatile tool for neuroimaging. Future work will focus on developing fully volumetric SSL architectures and multimodal foundation models that integrate imaging and textual data, alongside clinical validation to realise their full potential in precision diagnostics and foundational medical AI.

\section{Author contributions}

M.M.: Conceptualisation, Methodology, Data Curation, Model Development, Validation, Formal Analysis, Visualisation, Investigation, and Writing Original Draft. G.J.M.P.: Conceptualisation, Writing Review \& Editing, and Supervision. D.C.A.: Conceptualisation, Writing Review \& Editing, Supervision, Project Administration, and Funding Acquisition.

\section{Acknowledgements}
M.M. is supported by the Wellcome Trust (221915). G.J.M.P. is funded by the Alzheimer’s Society Heather Corrie Impact Fund (grant number 577 [AS-PG-21-045]). D.C.A. is supported by the Wellcome Trust (221915), EPSRC (EP/Y028856/1), and the NIHR UCLH Biomedical Research Centre.

The NACC database is funded by NIA/NIH Grant U24 AG072122. NACC data are contributed by the NIA-funded ADRCs: P30 AG062429 (PI James Brewer, MD, PhD), P30 AG066468 (PI Oscar Lopez, MD), P30 AG062421 (PI Bradley Hyman, MD, PhD), P30 AG066509 (PI Thomas Grabowski, MD), P30 AG066514 (PI Mary Sano, PhD), P30 AG066530 (PI Helena Chui, MD), P30 AG066507 (PI Marilyn Albert, PhD), P30 AG066444 (PI David Holtzman, MD), P30 AG066518 (PI Lisa Silbert, MD, MCR), P30 AG066512 (PI Thomas Wisniewski, MD), P30 AG066462 (PI Scott Small, MD), P30 AG072979 (PI David Wolk, MD), P30 AG072972 (PI Charles DeCarli, MD), P30 AG072976 (PI Andrew Saykin, PsyD), P30 AG072975 (PI Julie A. Schneider, MD, MS), P30 AG072978 (PI Ann McKee, MD), P30 AG072977 (PI Robert Vassar, PhD), P30 AG066519 (PI Frank LaFerla, PhD), P30 AG062677 (PI Ronald Petersen, MD, PhD), P30 AG079280 (PI Jessica Langbaum, PhD), P30 AG062422 (PI Gil Rabinovici, MD), P30 AG066511 (PI Allan Levey, MD, PhD), P30 AG072946 (PI Linda Van Eldik, PhD), P30 AG062715 (PI Sanjay Asthana, MD, FRCP), P30 AG072973 (PI Russell Swerdlow, MD), P30 AG066506 (PI Glenn Smith, PhD, ABPP), P30 AG066508 (PI Stephen Strittmatter, MD, PhD), P30 AG066515 (PI Victor Henderson, MD, MS), P30 AG072947 (PI Suzanne Craft, PhD), P30 AG072931 (PI Henry Paulson, MD, PhD), P30 AG066546 (PI Sudha Seshadri, MD), P30 AG086401 (PI Erik Roberson, MD, PhD), P30 AG086404 (PI Gary Rosenberg, MD), P20 AG068082 (PI Angela Jefferson, PhD), P30 AG072958 (PI Heather Whitson, MD), P30 AG072959 (PI James Leverenz, MD).

We thank all the institutions and consortia that made their datasets publicly available, enabling this research. We acknowledge the investigators and participants involved in the Alzheimer's Disease Neuroimaging Initiative (ADNI), the National Alzheimer’s Coordinating Center (NACC), and other contributing studies.

We are also grateful to the teams behind BraTS, ReMIND, UPENN-GBM, UCSF-PDGM, Burdenko-GBM-Progression, HABS, OASIS, AIBL, NIFD, and FeTA for their efforts in collecting and sharing high-quality data.

We gratefully acknowledge the data providers, funding sources, and individuals whose participation made this research possible.

\section{Competing interests}
All authors declare no financial or non-financial competing interests. 

\section{Data availability}

All datasets used in this study are publicly available and were accessed in compliance with their respective data use agreements:

\begin{itemize}
    \item \textbf{BraTS}~\cite{baid2021rsna}: The Brain Tumour Segmentation (BraTS) dataset is available at \url{https://www.kaggle.com/datasets/dschettler8845/brats-2021-task1}.
    
    \item \textbf{ReMIND}~\cite{juvekar2024remind}: Retrospective Multicenter Imaging of Neurodegeneration Dataset. Available at \url{https://www.cancerimagingarchive.net/collection/remind/}.
    
    \item \textbf{UPENN-GBM}~\cite{bakas2022university}: University of Pennsylvania Glioblastoma Dataset. Available at \url{https://www.cancerimagingarchive.net/collection/upenn-gbm/}.
    
    \item \textbf{UCSF-PDGM}~\cite{calabrese2022university}: UCSF Pediatric Diffuse Glioma MRI dataset. Available at \url{https://www.cancerimagingarchive.net/collection/ucsf-pdgm/}.
    
    \item \textbf{Burdenko-GBM-Progression}~\cite{zolotova2023burdenko}: Burdenko Neurosurgical Institute GBM Progression dataset. Available at \url{https://www.cancerimagingarchive.net/collection/burdenko-gbm-progression/}.
    
    \item \textbf{HABS}~\cite{dagley2017harvard}: Harvard Aging Brain Study. Available at \url{https://habs.mgh.harvard.edu/}.
    
    \item \textbf{OASIS}~\cite{lamontagne2019oasis}: A Longitudinal Study of Normal Aging and Alzheimer's Disease. Available at \url{http://www.oasis-brains.org}.
    
    \item \textbf{ADNI}~\cite{petersen2010alzheimer}: Alzheimer's Disease Neuroimaging Initiative. Available at \url{http://adni.loni.usc.edu}.
    
    \item \textbf{NACC}~\cite{beekly2007national}: National Alzheimer's Coordinating Center. Data contributed by NIA-funded ADRCs; see \url{https://naccdata.org}.
    
    \item \textbf{AIBL}~\cite{ellis2009australian}: Australian Imaging, Biomarker \& Lifestyle Flagship Study of Ageing. Available at \url{https://aibl.csiro.au}.
    
    \item \textbf{NIFD}~\cite{rohrer2013neuroimaging}: Neuroimaging Initiative for Frontotemporal Dementia. Available at \url{https://ida.loni.usc.edu/login.jsp?project=NIFD}.
    
    \item \textbf{FeTA}~\cite{payette2021fetal}: Fetal Brain MRI Atlas. Available at \url{https://feta.grand-challenge.org/feta-2022/}.
\end{itemize}

All data were used in accordance with relevant institutional and ethical guidelines. No personally identifiable information was accessed in this study.

\section{Code availability}
The code used to train, fine-tune, and evaluate BrainFound is available at \url{https://github.com/Moona-Mazher/BrainFound}, which is based on PyTorch. Brain MRI images were processed using the MRIPreprocessor tool (\url{https://github.com/ReubenDo/MRIPreprocessor}). Image data were sourced in both NIfTI and DICOM formats; DICOM files were extracted using Pydicom v2.3.0. Reading and loading of 3D MRI images were performed using SimpleITK v2.2.1 and Nibabel v3.2.1. Results were analyzed and visualized with Python v3.6, using NumPy v1.19.5, SciPy v1.5.4, seaborn v0.12.0, Matplotlib v3.6.1, pandas v1.5.0, Scikit-Learn v1.1.3, and Pillow v9.2.0. Additional result analysis was performed using 3D-Slicer v5.4.1 and ITK-Snap v3.8.0. 

\begin{appendices}




\end{appendices}


\bibliography{sn-bibliography}

@article{hussain2022modern,
  title={Modern diagnostic imaging technique applications and risk factors in the medical field: a review},
  author={Hussain, Shah and Mubeen, Iqra and Ullah, Niamat and Shah, Syed Shahab Ud Din and Khan, Bakhtawar Abduljalil and Zahoor, Muhammad and Ullah, Riaz and Khan, Farhat Ali and Sultan, Mujeeb A},
  journal={BioMed research international},
  volume={2022},
  number={1},
  pages={5164970},
  year={2022},
  publisher={Wiley Online Library}
}

@article{loginoff2023advancements,
  title={Advancements in radiology and diagnostic imaging},
  author={{\L}oginoff, Jan and Augustynowicz, Kinga and {\'S}wi{\k{a}}der, Kinga and Ostaszewska, Sandra and Morawski, Przemys{\l}aw and Pactwa, Filip and Popi{\'n}ska, Zuzanna},
  journal={Journal of Education, Health and Sport},
  volume={33},
  number={1},
  pages={45--51},
  year={2023}
}

@article{yoon2024use,
  title={Use of artificial intelligence in triaging of chest radiographs to reduce radiologists’ workload},
  author={Yoon, Sung Hyun and Park, Sunyoung and Jang, Sowon and Kim, Junghoon and Lee, Kyung Won and Lee, Woojoo and Lee, Seungjae and Yun, Gabin and Lee, Kyung Hee},
  journal={European Radiology},
  volume={34},
  number={2},
  pages={1094--1103},
  year={2024},
  publisher={Springer}
}

@article{kwee2021workload,
  title={Workload of diagnostic radiologists in the foreseeable future based on recent scientific advances: growth expectations and role of artificial intelligence},
  author={Kwee, Thomas C and Kwee, Robert M},
  journal={Insights into imaging},
  volume={12},
  pages={1--12},
  year={2021},
  publisher={Springer}
}

@article{kasalak2023work,
  title={Work overload and diagnostic errors in radiology},
  author={Kasalak, {\"O}mer and Alnahwi, Haider and Toxopeus, Romy and Pennings, Jan P and Yakar, Derya and Kwee, Thomas C},
  journal={European Journal of Radiology},
  volume={167},
  pages={111032},
  year={2023},
  publisher={Elsevier}
}

@article{pesapane2024errors,
  title={Errors in Radiology: A Standard Review},
  author={Pesapane, Filippo and Gnocchi, Giulia and Quarrella, Cettina and Sorce, Adriana and Nicosia, Luca and Mariano, Luciano and Bozzini, Anna Carla and Marinucci, Irene and Priolo, Francesca and Abbate, Francesca and others},
  journal={Journal of Clinical Medicine},
  volume={13},
  number={15},
  pages={4306},
  year={2024}
}

@article{litjens2017survey,
  title={A survey on deep learning in medical image analysis},
  author={Litjens, Geert and Kooi, Thijs and Bejnordi, Babak Ehteshami and Setio, Arnaud Arindra Adiyoso and Ciompi, Francesco and Ghafoorian, Mohsen and Van Der Laak, Jeroen Awm and Van Ginneken, Bram and S{\'a}nchez, Clara I},
  journal={Medical image analysis},
  volume={42},
  pages={60--88},
  year={2017},
  publisher={Elsevier}
}

@article{galbusera2024image,
  title={Image annotation and curation in radiology: an overview for machine learning practitioners},
  author={Galbusera, Fabio and Cina, Andrea},
  journal={European Radiology Experimental},
  volume={8},
  number={1},
  pages={11},
  year={2024},
  publisher={Springer}
}

@article{yoon2024domain,
  title={Domain Generalization for Medical Image Analysis: A Review},
  author={Yoon, Jee Seok and Oh, Kwanseok and Shin, Yooseung and Mazurowski, Maciej A and Suk, Heung-Il},
  journal={Proceedings of the IEEE},
  year={2024},
  publisher={IEEE}
}

@article{zhang2023dive,
  title={Dive into the details of self-supervised learning for medical image analysis},
  author={Zhang, Chuyan and Zheng, Hao and Gu, Yun},
  journal={Medical Image Analysis},
  volume={89},
  pages={102879},
  year={2023},
  publisher={Elsevier}
}

@article{schneider2024foundation,
  title={Foundation models: a new paradigm for artificial intelligence},
  author={Schneider, Johannes and Meske, Christian and Kuss, Pauline},
  journal={Business \& Information Systems Engineering},
  volume={66},
  number={2},
  pages={221--231},
  year={2024},
  publisher={Springer}
}

@article{brown2020language,
  title={Language models are few-shot learners},
  author={Brown, Tom and Mann, Benjamin and Ryder, Nick and Subbiah, Melanie and Kaplan, Jared D and Dhariwal, Prafulla and Neelakantan, Arvind and Shyam, Pranav and Sastry, Girish and Askell, Amanda and others},
  journal={Advances in neural information processing systems},
  volume={33},
  pages={1877--1901},
  year={2020}
}

@inproceedings{hosseinzadeh2023towards,
  title={Towards foundation models learned from anatomy in medical imaging via self-supervision},
  author={Hosseinzadeh Taher, Mohammad Reza and Gotway, Michael B and Liang, Jianming},
  booktitle={MICCAI Workshop on Domain Adaptation and Representation Transfer},
  pages={94--104},
  year={2023},
  organization={Springer}
}

@inproceedings{shivdeo2021evaluation,
  title={Evaluation of 3d and 2d deep learning techniques for semantic segmentation in ct scans},
  author={Shivdeo, Abhishek and Lokwani, Rohit and Kulkarni, Viraj and Kharat, Amit and Pant, Aniruddha},
  booktitle={2021 International Conference on Artificial Intelligence, Big Data, Computing and Data Communication Systems (icABCD)},
  pages={1--8},
  year={2021},
  organization={IEEE}
}

@article{ferraro2017multimodal,
  title={Multimodal structural MRI in the diagnosis of motor neuron diseases},
  author={Ferraro, Pilar M and Agosta, Federica and Riva, Nilo and Copetti, Massimiliano and Spinelli, Edoardo Gioele and Falzone, Yuri and Sorar{\`u}, Gianni and Comi, Giancarlo and Chi{\`o}, Adriano and Filippi, Massimo},
  journal={NeuroImage: Clinical},
  volume={16},
  pages={240--247},
  year={2017},
  publisher={Elsevier}
}

@article{weiner2013alzheimer,
  title={The Alzheimer's Disease Neuroimaging Initiative: a review of papers published since its inception},
  author={Weiner, Michael W and Veitch, Dallas P and Aisen, Paul S and Beckett, Laurel A and Cairns, Nigel J and Green, Robert C and Harvey, Danielle and Jack, Clifford R and Jagust, William and Liu, Enchi and others},
  journal={Alzheimer's \& Dementia},
  volume={9},
  number={5},
  pages={e111--e194},
  year={2013},
  publisher={Elsevier}
}

@article{jiang2023self,
  title={Self-supervised learning for early detection of neurodegenerative diseases with small data},
  author={Jiang, Hongchao},
  year={2023},
  publisher={Nanyang Technological University}
}

@article{zhang2016multimodal,
  title={Multimodal MRI-based classification of trauma survivors with and without post-traumatic stress disorder},
  author={Zhang, Qiongmin and Wu, Qizhu and Zhu, Hongru and He, Ling and Huang, Hua and Zhang, Junran and Zhang, Wei},
  journal={Frontiers in neuroscience},
  volume={10},
  pages={292},
  year={2016},
  publisher={Frontiers Media SA}
}

@article{zhao2021longitudinal,
  title={Longitudinal self-supervised learning},
  author={Zhao, Qingyu and Liu, Zixuan and Adeli, Ehsan and Pohl, Kilian M},
  journal={Medical image analysis},
  volume={71},
  pages={102051},
  year={2021},
  publisher={Elsevier}
}

@inproceedings{chen2020simple,
  title={A simple framework for contrastive learning of visual representations},
  author={Chen, Ting and Kornblith, Simon and Norouzi, Mohammad and Hinton, Geoffrey},
  booktitle={International conference on machine learning},
  pages={1597--1607},
  year={2020},
  organization={PmLR}
}

@inproceedings{chen2021empirical,
  title={An empirical study of training self-supervised vision transformers},
  author={Chen, Xinlei and Xie, Saining and He, Kaiming},
  booktitle={Proceedings of the IEEE/CVF international conference on computer vision},
  pages={9640--9649},
  year={2021}
}

@inproceedings{he2022masked,
  title={Masked autoencoders are scalable vision learners},
  author={He, Kaiming and Chen, Xinlei and Xie, Saining and Li, Yanghao and Doll{\'a}r, Piotr and Girshick, Ross},
  booktitle={Proceedings of the IEEE/CVF conference on computer vision and pattern recognition},
  pages={16000--16009},
  year={2022}
}

@article{oquab2023dinov2,
  title={Dinov2: Learning robust visual features without supervision},
  author={Oquab, Maxime and Darcet, Timoth{\'e}e and Moutakanni, Th{\'e}o and Vo, Huy and Szafraniec, Marc and Khalidov, Vasil and Fernandez, Pierre and Haziza, Daniel and Massa, Francisco and El-Nouby, Alaaeldin and others},
  journal={arXiv preprint arXiv:2304.07193},
  year={2023}
}

@article{baid2021rsna,
  title={The rsna-asnr-miccai brats 2021 benchmark on brain tumor segmentation and radiogenomic classification},
  author={Baid, Ujjwal and Ghodasara, Satyam and Mohan, Suyash and Bilello, Michel and Calabrese, Evan and Colak, Errol and Farahani, Keyvan and Kalpathy-Cramer, Jayashree and Kitamura, Felipe C and Pati, Sarthak and others},
  journal={arXiv preprint arXiv:2107.02314},
  year={2021}
}

@article{juvekar2024remind,
  title={Remind: The brain resection multimodal imaging database},
  author={Juvekar, Parikshit and Dorent, Reuben and K{\"o}gl, Fryderyk and Torio, Erickson and Barr, Colton and Rigolo, Laura and Galvin, Colin and Jowkar, Nick and Kazi, Anees and Haouchine, Nazim and others},
  journal={Scientific Data},
  volume={11},
  number={1},
  pages={494},
  year={2024},
  publisher={Nature Publishing Group UK London}
}

@article{bakas2022university,
  title={The University of Pennsylvania glioblastoma (UPenn-GBM) cohort: advanced MRI, clinical, genomics, \& radiomics},
  author={Bakas, Spyridon and Sako, Chiharu and Akbari, Hamed and Bilello, Michel and Sotiras, Aristeidis and Shukla, Gaurav and Rudie, Jeffrey D and Santamar{\'\i}a, Natali Flores and Kazerooni, Anahita Fathi and Pati, Sarthak and others},
  journal={Scientific data},
  volume={9},
  number={1},
  pages={453},
  year={2022},
  publisher={Nature Publishing Group UK London}
}

@article{calabrese2022university,
  title={The University of California San Francisco preoperative diffuse glioma MRI dataset},
  author={Calabrese, Evan and Villanueva-Meyer, Javier E and Rudie, Jeffrey D and Rauschecker, Andreas M and Baid, Ujjwal and Bakas, Spyridon and Cha, Soonmee and Mongan, John T and Hess, Christopher P},
  journal={Radiology: Artificial Intelligence},
  volume={4},
  number={6},
  pages={e220058},
  year={2022},
  publisher={Radiological Society of North America}
}

@article{zolotova2023burdenko,
  title={Burdenko’s glioblastoma progression dataset (Burdenko-GBM-progression)(Version 1)[Data set]},
  author={Zolotova, SV and Golanov, AV and Pronin, IN and Dalechina, AV and Nikolaeva, AA and Belyashova, AS and Usachev, DY and Kondrateva, EA and Druzhinina, PV and Shirokikh, BN and others},
  journal={The Cancer Imaging Archive},
  year={2023}
}

@article{dagley2017harvard,
  title={Harvard aging brain study: dataset and accessibility},
  author={Dagley, Alexander and LaPoint, Molly and Huijbers, Willem and Hedden, Trey and McLaren, Donald G and Chatwal, Jasmeer P and Papp, Kathryn V and Amariglio, Rebecca E and Blacker, Deborah and Rentz, Dorene M and others},
  journal={Neuroimage},
  volume={144},
  pages={255--258},
  year={2017},
  publisher={Elsevier}
}

@article{lamontagne2019oasis,
  title={OASIS-3: longitudinal neuroimaging, clinical, and cognitive dataset for normal aging and Alzheimer disease},
  author={LaMontagne, Pamela J and Benzinger, Tammie LS and Morris, John C and Keefe, Sarah and Hornbeck, Russ and Xiong, Chengjie and Grant, Elizabeth and Hassenstab, Jason and Moulder, Krista and Vlassenko, Andrei G and others},
  journal={medrxiv},
  pages={2019--12},
  year={2019},
  publisher={Cold Spring Harbor Laboratory Press}
}

@article{petersen2010alzheimer,
  title={Alzheimer's disease Neuroimaging Initiative (ADNI) clinical characterization},
  author={Petersen, Ronald Carl and Aisen, Paul S and Beckett, Laurel A and Donohue, Michael C and Gamst, Anthony Collins and Harvey, Danielle J and Jack Jr, CR and Jagust, William J and Shaw, Leslie M and Toga, Arthur W and others},
  journal={Neurology},
  volume={74},
  number={3},
  pages={201--209},
  year={2010},
  publisher={Lippincott Williams \& Wilkins}
}

@article{beekly2007national,
  title={The National Alzheimer's Coordinating Center (NACC) database: the uniform data set},
  author={Beekly, Duane L and Ramos, Erin M and Lee, William W and Deitrich, Woodrow D and Jacka, Mary E and Wu, Joylee and Hubbard, Janene L and Koepsell, Thomas D and Morris, John C and Kukull, Walter A and others},
  journal={Alzheimer Disease \& Associated Disorders},
  volume={21},
  number={3},
  pages={249--258},
  year={2007},
  publisher={LWW}
}

@article{ellis2009australian,
  title={The Australian Imaging, Biomarkers and Lifestyle (AIBL) study of aging: methodology and baseline characteristics of 1112 individuals recruited for a longitudinal study of Alzheimer's disease},
  author={Ellis, Kathryn A and Bush, Ashley I and Darby, David and De Fazio, Daniela and Foster, Jonathan and Hudson, Peter and Lautenschlager, Nicola T and Lenzo, Nat and Martins, Ralph N and Maruff, Paul and others},
  journal={International psychogeriatrics},
  volume={21},
  number={4},
  pages={672--687},
  year={2009},
  publisher={Cambridge University Press}
}

@article{rohrer2013neuroimaging,
  title={Neuroimaging in frontotemporal dementia},
  author={Rohrer, Jonathan D and Rosen, Howard J},
  journal={International Review of Psychiatry},
  volume={25},
  number={2},
  pages={221--229},
  year={2013},
  publisher={Taylor \& Francis}
}

@inproceedings{payette2021fetal,
  title={Fetal Brain Tissue Annotation and Segmentation Challenge},
  author={Payette, Kelly and Dumast, Priscille de and Jakab, Andras and Cuadra, Meritxell Bach and Vasung, Lana and Licandro, Roxane and Menze, Bjoern and Li, Hongwei},
  booktitle={24th International Conference on Medical Image Computing and Computer Assisted Intervention (MICCAI 2021)},
  pages={15},
  year={2021},
  organization={Zenodo}
}

@article{karnyoto2024swav,
  title={SwAV transfer learning and knowledge distillation on chest X-ray classification},
  author={Karnyoto, Andrea Stevens and Elwirehardja, Gregorius Natanael and Cenggoro, Tjeng Wawan and Pardamean, Bens and others},
  journal={Commun. Math. Biol. Neurosci.},
  volume={2024},
  pages={Article--ID},
  year={2024}
}

@article{ma2024segment,
  title={Segment anything in medical images},
  author={Ma, Jun and He, Yuting and Li, Feifei and Han, Lin and You, Chenyu and Wang, Bo},
  journal={Nature Communications},
  volume={15},
  number={1},
  pages={654},
  year={2024},
  publisher={Nature Publishing Group UK London}
}

@article{tang2025hi,
  title={Hi-End-MAE: Hierarchical encoder-driven masked autoencoders are stronger vision learners for medical image segmentation},
  author={Tang, Fenghe and Yao, Qingsong and Ma, Wenxin and Wu, Chenxu and Jiang, Zihang and Zhou, S Kevin},
  journal={arXiv preprint arXiv:2502.08347},
  year={2025}
}

@article{MONAI2022,
  title = {MONAI: An open-source framework for deep learning in healthcare},
  author = {The MONAI Consortium},
  journal = {arXiv preprint arXiv:2211.02701},
  year = {2022},
  url = {https://arxiv.org/abs/2211.02701}
}

@article{raghu2019transfusion,
  title={Transfusion: Understanding transfer learning for medical imaging},
  author={Raghu, Maithra and Zhang, Chiyuan and Kleinberg, Jon and Bengio, Samy},
  journal={Advances in neural information processing systems},
  volume={32},
  year={2019}
}

@article{yamashita2018convolutional,
  title={Convolutional neural networks: an overview and application in radiology},
  author={Yamashita, Rikiya and Nishio, Mizuho and Do, Richard Kinh Gian and Togashi, Kaori},
  journal={Insights into imaging},
  volume={9},
  number={4},
  pages={611--629},
  year={2018},
  publisher={Springer}
}

@article{zhou2023foundation,
  title={A foundation model for generalizable disease detection from retinal images},
  author={Zhou, Yukun and Chia, Mark A and Wagner, Siegfried K and Ayhan, Murat S and Williamson, Dominic J and Struyven, Robbert R and Liu, Timing and Xu, Moucheng and Lozano, Mateo G and Woodward-Court, Peter and others},
  journal={Nature},
  volume={622},
  number={7981},
  pages={156--163},
  year={2023},
  publisher={Nature Publishing Group UK London}
}

@article{baharoon2023evaluating,
  title={Evaluating general purpose vision foundation models for medical image analysis: An experimental study of dinov2 on radiology benchmarks},
  author={Baharoon, Mohammed and Qureshi, Waseem and Ouyang, Jiahong and Xu, Yanwu and Aljouie, Abdulrhman and Peng, Wei},
  journal={arXiv preprint arXiv:2312.02366},
  year={2023}
}

@article{qayyum2025foundation,
  title={Foundation Model for Whole-Heart Segmentation: Leveraging Student-Teacher Learning in Multi-Modal Medical Imaging},
  author={Qayyum, Abdul and Mazher, Moona and Ugurlu, Devran and Lemus, Jose Alonso Solis and Rodero, Cristobal and Niederer, Steven A},
  journal={arXiv preprint arXiv:2503.19005},
  year={2025}
}

@article{dorent2021learning,
  title={Learning joint segmentation of tissues and brain lesions from task-specific hetero-modal domain-shifted datasets},
  author={Dorent, Reuben and Booth, Thomas and Li, Wenqi and Sudre, Carole H and Kafiabadi, Sina and Cardoso, Jorge and Ourselin, Sebastien and Vercauteren, Tom},
  journal={Medical image analysis},
  volume={67},
  pages={101862},
  year={2021},
  publisher={Elsevier}
}

\appendix

\section{BrainFound Pseudocode}

\begin{figure}[H]
    \centering
    \includegraphics[width=1\textwidth]{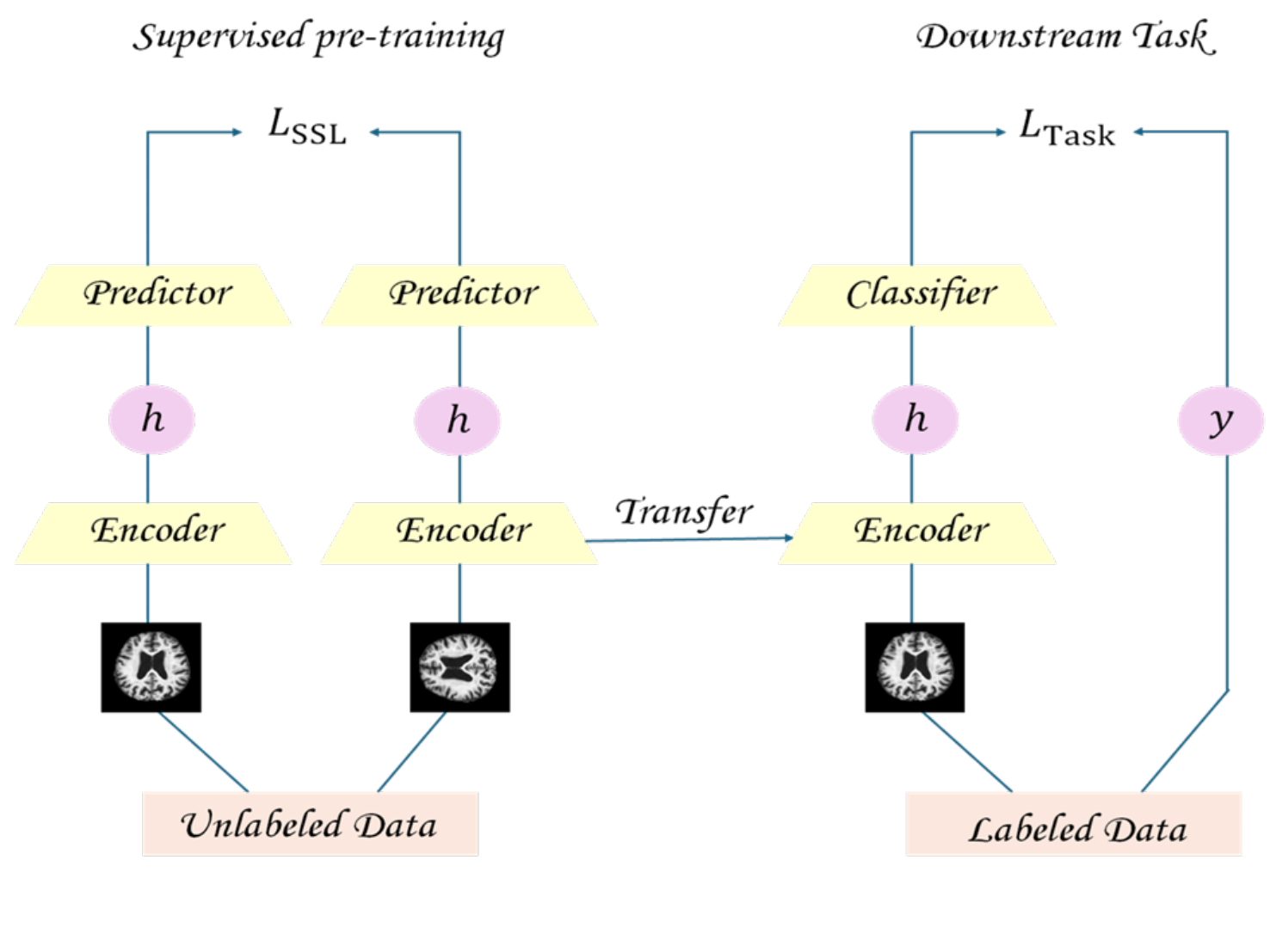}
    \caption{Proposed architecture of BrainFound adapted from DINOv2 \cite{oquab2023dinov2}.
}
    \label{fig:dinov2}
\end{figure}

\begin{figure}[H]
    \centering
    \includegraphics[width=1\textwidth]{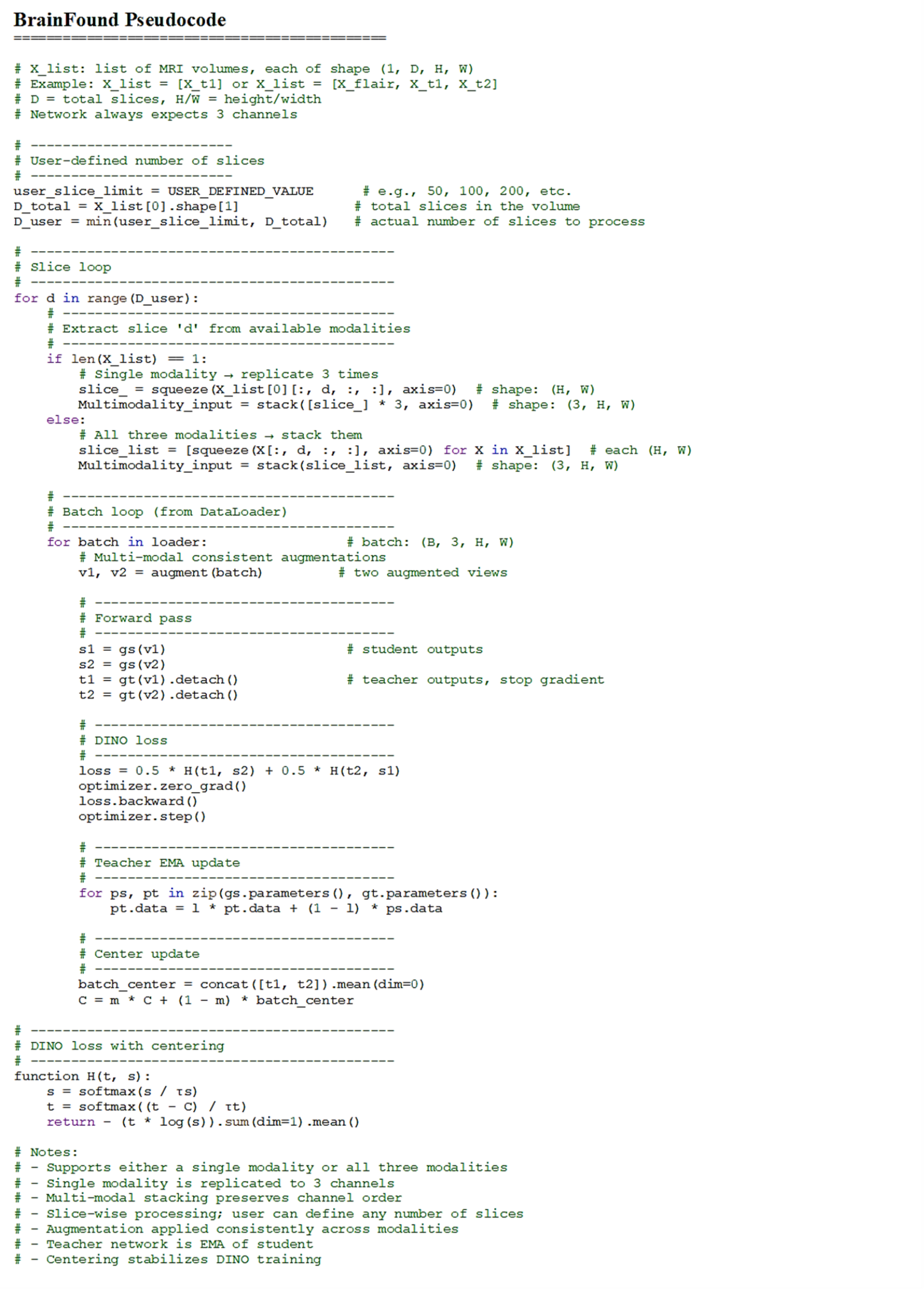}
    \caption{Pseudocode for BrainFound.
}
    \label{fig:pseudocode}
\end{figure}

\section{Dataset distribution for SSL pretraining and downstream task}

\begin{table}[h]
\centering
\caption{Dataset distribution for SSL pretraining and downstream tasks. HABS, ReMIND, UPENN-GBM, and Burdenko-GBM-Progression were used only for SSL pretraining.}
\label{tab:dataset_distribution}
\begin{tabular}{l r r r}
\toprule
\textbf{Dataset / Source} & \textbf{Total Images} & \textbf{Training / Validation} & \textbf{Testing} \\
\midrule
NACC \cite{beekly2007national} & 1,812 & 1,540 & 272 \\
OASIS \cite{lamontagne2019oasis} & 4,134 & 3,514 & 620 \\
HABS \cite{dagley2017harvard} & 580 & - & - \\
ADNI \cite{petersen2010alzheimer} & 382 & 235 & 57 \\
AIBL \cite{ellis2009australian} & 152 & 129 & 23 \\
NIFD \cite{rohrer2013neuroimaging} & 346 & 294 & 52 \\
BraTS \cite{baid2021rsna} & 1,107 & 941 & 166 \\
FeTA \cite{payette2021fetal} & 80 & 68 & 12 \\
ReMIND \cite{juvekar2024remind} & 114 & - & - \\
UPENN-GBM \cite{bakas2022university} & 1,008 & - & - \\
UCSF-PDGM \cite{calabrese2022university} & 588 & 500 & 88 \\
Burdenko-GBM-Progression \cite{zolotova2023burdenko} & 360 & - & - \\
\midrule
\textbf{Total} & \textbf{10,663} & \textbf{9,286} & \textbf{1,377} \\
\bottomrule
\end{tabular}
\end{table}

\section{Comparison of single (T1, T2, and Flair) vs combined modalities}

\begin{figure}[H]
    \centering
    \includegraphics[width=1\textwidth]{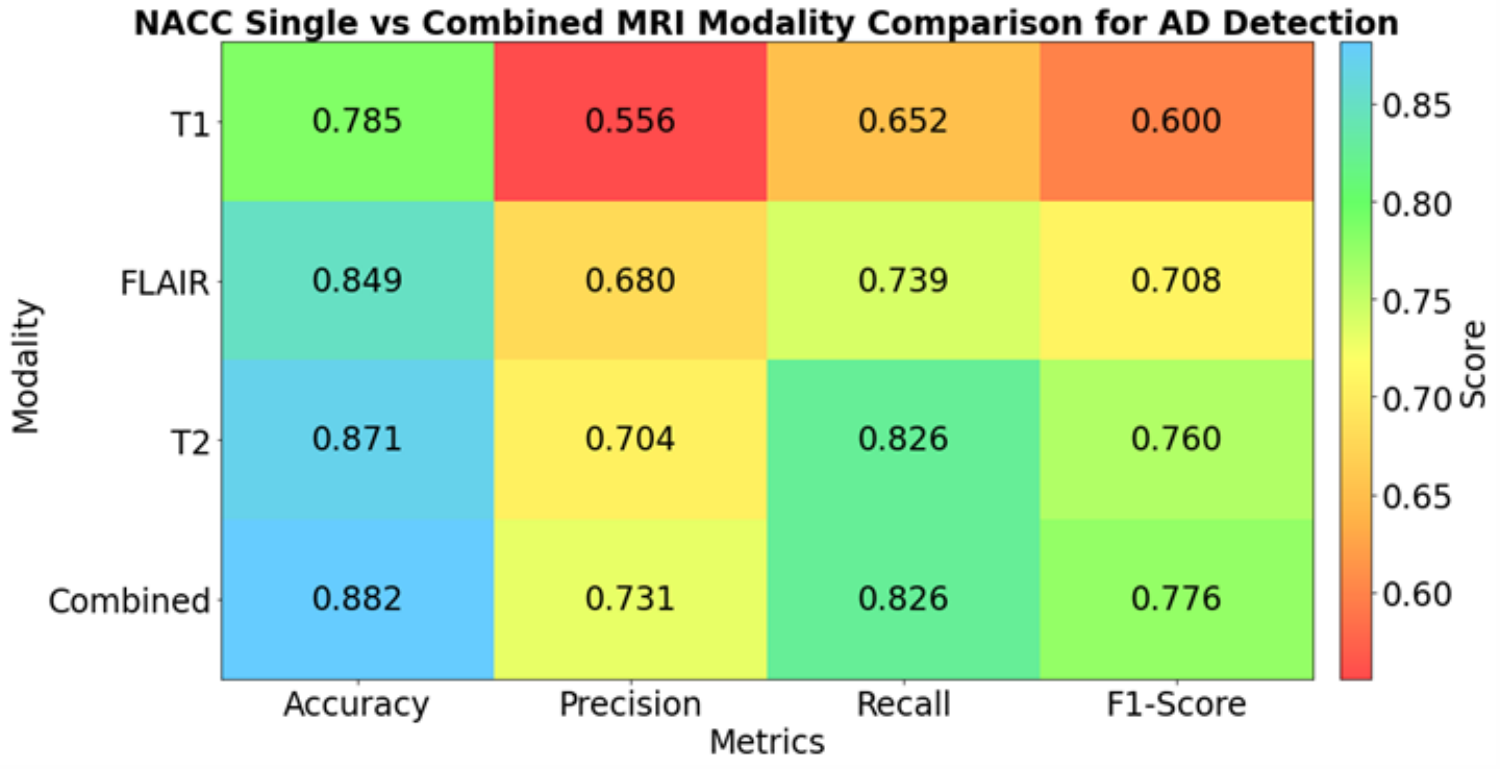}
    \caption{Comparison of single (T1, T2, and Flair) vs combined modalities on NACC dataset for AD detection.
}
    \label{fig:singlevsmulti}
\end{figure}

The figure compares Alzheimer’s disease detection performance on the NACC cohort using single-modality MRI inputs (T1, T2, and FLAIR) versus a combined multimodal setting. Across all evaluation metrics—accuracy, precision, recall, and F1-score—the combined-modality configuration consistently achieves the highest or near-highest performance, indicating the benefit of integrating complementary MRI contrasts.

Among single modalities, T2-weighted images generally outperform T1 and FLAIR, particularly in accuracy and recall, suggesting stronger sensitivity to disease-related structural changes. However, no single modality achieves optimal performance across all metrics. In contrast, the combined multimodal approach yields the highest overall accuracy (0.882) and F1-score (0.776), while maintaining strong precision and recall, demonstrating a more balanced and robust classification performance.

These results highlight that multimodal fusion enables the model to capture complementary anatomical and pathological information present across different MRI sequences, leading to improved discriminative representations and more reliable disease detection compared with single-modality inputs.
\end{document}